%% file: main.tex
\documentclass[lettersize,journal]{IEEEtran}
\usepackage{amsmath,amsfonts}
\usepackage{algorithmic}
\usepackage{algorithm}
\usepackage{array}
\usepackage[caption=false,font=normalsize,labelfont=rm,textfont=rm]{subfig}
\usepackage{textcomp}
\usepackage{stfloats}
\usepackage{url}
\usepackage{verbatim}
\usepackage{graphicx}
\usepackage{cite}
\usepackage{booktabs} 
\usepackage{xcolor}
\usepackage{multirow}
\usepackage{pifont}
\usepackage[normalem]{ulem}
\usepackage[
  colorlinks=true,       
  linkcolor=red,         
  citecolor=green,        
  filecolor=magenta,    
  urlcolor=cyan,         
  bookmarks=true,        
  pdfborder={0 0 0}      
]{hyperref}
\begin{document}

\title{Memory-Augmented Query Intent Understanding for Efficient Chat-based Image Retrieval}

\author{Xianke~Chen, Daizong~Liu,  Yushuo~Lou,
        Xin~Tan, Xun~Yang, \\  Shuhui~Wang,  Xun~Wang,~\IEEEmembership{Member,~IEEE}, Jianfeng~Dong
\IEEEcompsocitemizethanks{
\IEEEcompsocthanksitem X. Chen is with the School of Computer Science and Technology, and the School of Statistics and Mathematics, Zhejiang Gongshang University, Hangzhou 310035, China. E-mail: cxkxk\_@outlook.com 
\IEEEcompsocthanksitem J. Dong and X. Wang  are with the the School of Computer Science and Technology, Zhejiang Gongshang University, and the Zhejiang Key Laboratory of Big Data and Future E-Commerce Technology,  Hangzhou 310035, China. 
\protect
E-mail: dongjf24@gmail.com,  wx@zjgsu.edu.cn
\IEEEcompsocthanksitem D. Liu is with the Wangxuan Institute of Computer Technology, Peking University, No. 128, Zhongguancun North Street, Beijing 100871, China. 
\protect
E-mail: dzliu@stu.pku.edu.cn
\IEEEcompsocthanksitem Y. Lou is with the School of Information and Electronic Engineering, Zhejiang Gongshang University, Hangzhou 310035, China. E-mail: 2336010117@pop.zjgsu.edu.cn
\IEEEcompsocthanksitem X. Tan is with the School of Computer Science and Technology, East China Normal University, Shanghai 200050, China. E-mail: xtan@cs.ecnu.edu.cn
\IEEEcompsocthanksitem S. Wang is with the Key Laboratory of Intelligent Information Processing of Chinese Academy of Sciences , Institute of Computing Technology, CAS, Beijing, 100190 China. E-mail: wangshuhui@ict.ac.cn
\IEEEcompsocthanksitem X. Yang is with the School of Information Science and Technology, University of Science and Technology of China, Hefei 230026, China.
\protect
E-mail: xyang21@ustc.edu.cn
}
\thanks{Corresponding author: Jianfeng Dong}}




\maketitle

\begin{abstract}
Different from traditional text-to-image retrieval tasks, chat-based image retrieval allows the human-interactive system to iteratively clarify and refine user intent through multi-round dialogue, thereby achieving more fine-grained retrieval results.
The key challenge in this task lies in dynamically understanding and updating the user's query intent across dialogue rounds.
Although existing works have achieved great performance on this new task, they simply handle history query information either by directly concatenating all previous queries into a long textual sequence or by relying on large language models to reconstruct the current query from history. Such strategies are computationally redundant and easily lead to inconsistent intent representations as the dialogue progresses. 
To alleviate these issues, this paper proposes a novel and efficient memory-based user intent updating framework for the chat-based image retrieval task, called Memory-Augmented Query Intent Understanding (MAQIU). It introduces a lightweight memorization module that dynamically aggregates and evolves the semantic representation of query intent across dialogues, while a memory recall mechanism is further employed to prevent intent forgetting and enhance long-term semantic integrity. In addition, MAQIU also integrates historical image retrieval results as visual guidance, allowing the model to strengthen cross-round correlations and refine current visual understanding.
Extensive experiments demonstrate that MAQIU achieves substantial performance gains while maintaining high computational efficiency, reducing dialogue encoding FLOPs by 86.4\% compared with the prior baseline ChatIR. Source code is available at \url{https://github.com/HuiGuanLab/MAQIU}.
\end{abstract}


\begin{figure}[!t]
\centering\includegraphics[width=0.98\columnwidth]{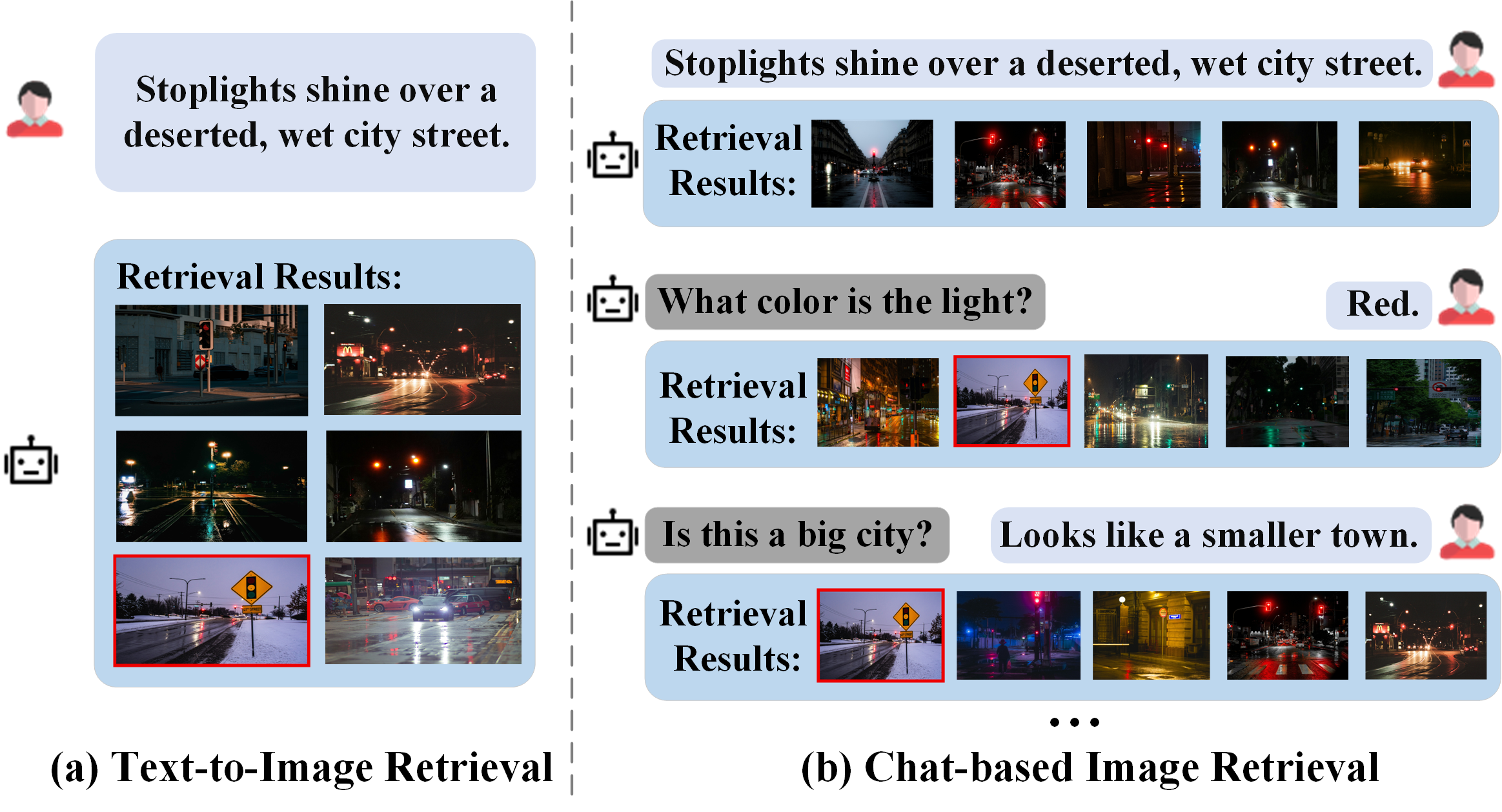}
\caption{The workflows of traditional text-to-image retrieval and chat-based image retrieval. (a) Traditional text-to-image retrieval searches images based solely on a single textual query. (b) Chat-based image retrieval conducts the search through multiple rounds of interaction, where the system asks questions and incorporates the user’s responses to progressively refine the query intent.}\label{fig:task}
\end{figure}

\section{Introduction}

\input{intro}

\section{Related work}
\input{rel_work}

\section{Method}
\input{method}

\input{experiment}

\bibliographystyle{IEEEtran}
\bibliography{IEEEabrv,reference}

\vspace{-10mm}

\begin{IEEEbiography}
[{\includegraphics[width=1in,height=1.25in,clip,keepaspectratio]{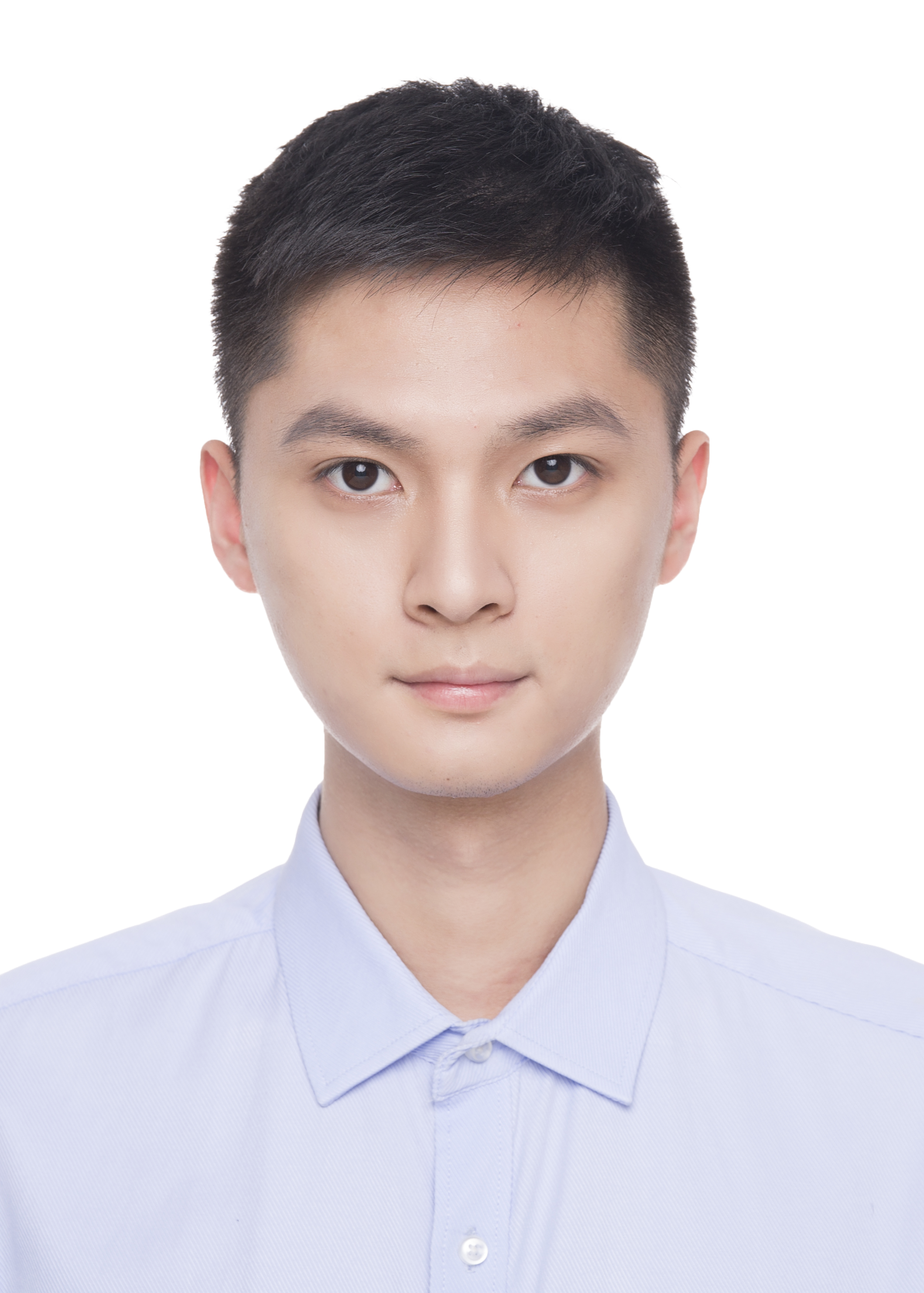}}]{Xianke Chen}
received the B.E. degree in network engineering from Zhejiang Gongshang University, Hangzhou, China, in 2020, and the M.E. degree from the College of Computer Science and Technology, Zhejiang Gongshang University, Hangzhou, China, in 2023. He is currently pursuing the Ph.D. degree with
the Big data statistics, Zhejiang Gongshang University. His research interests include multi-modal learning.
\end{IEEEbiography}


\begin{IEEEbiography}
[{\includegraphics[width=1in,height=1.25in,clip,keepaspectratio]{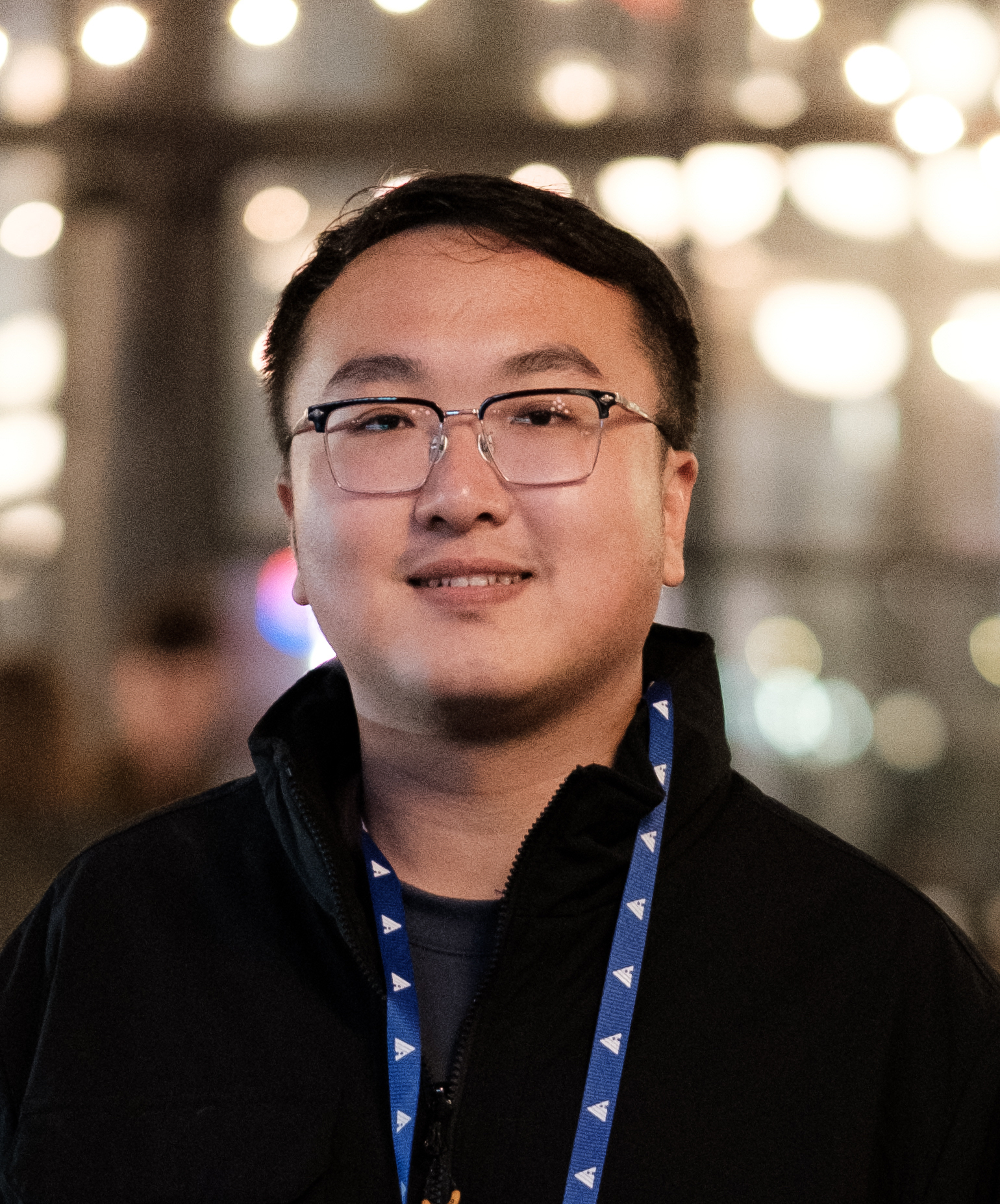}}]{Daizong Liu} received the M.S. degree in Electronic Information and Communication of Huazhong University of Science and Technology in 2021. He is currently working toward the Ph.D. degree at Wangxuan Institute of Computer Technology of Peking University. His research interests include 3D adversarial attacks, multi-modal learning, LVLM robustness, etc. He has published more than 40 papers in refereed conference proceedings and journals such as TPAMI, NeurIPS, CVPR, ICCV, ECCV, SIGIR, AAAI. He regularly serves on the program committees of top-tier AI conferences such as NeurIPS, ICML, ICLR, CVPR, ICCV and ACL.
\end{IEEEbiography}


\begin{IEEEbiography}
[{\includegraphics[width=1in,height=1.25in,clip,keepaspectratio]{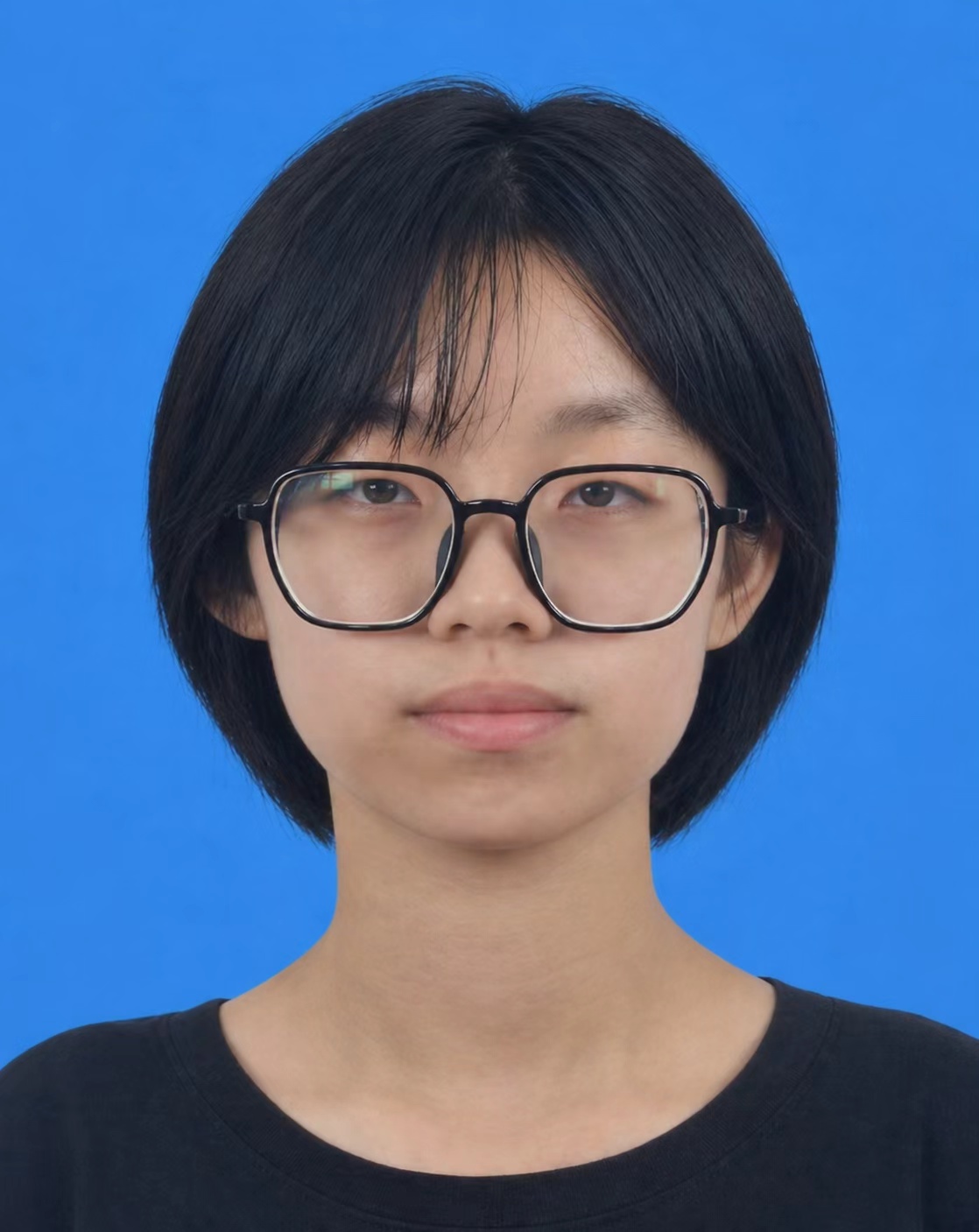}}]{Yushuo Lou}  is currently pursuing the B.E. degree in Artificial Intelligence with the School of Information and Electronic Engineering, Zhejiang Gongshang University, Hangzhou, China. Her research interests include multimodal learning.
\end{IEEEbiography}


\begin{IEEEbiography}
[{\includegraphics[width=1in,height=1.25in,clip,keepaspectratio]{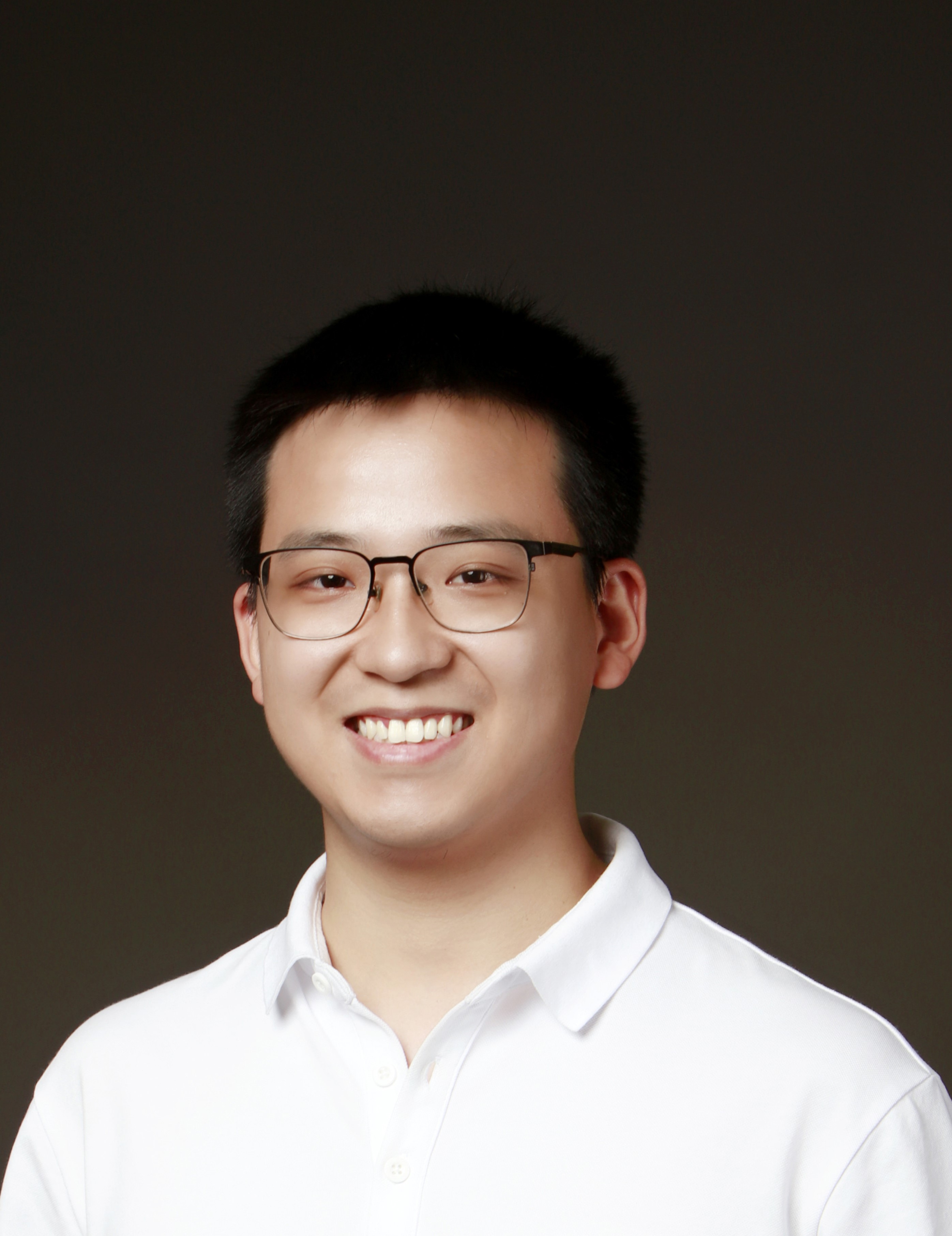}}]{Xin Tan} received the BE degree in automation from the Chongqing University, China in 2017 and the dual PhD degrees in computer science from Shanghai Jiao Tong University and the City University of Hong Kong. He is currently with East China
Normal University, China. His research interests
include in computer vision and deep learning. He
serves as a program committee member/reviewer
for CVPR, ICCV, AAAI, IJCAI and International
Journal of Computer Vision (IJCV).
\end{IEEEbiography}



\begin{IEEEbiography}[{\includegraphics[width=1in,height=1.25in,clip,keepaspectratio]{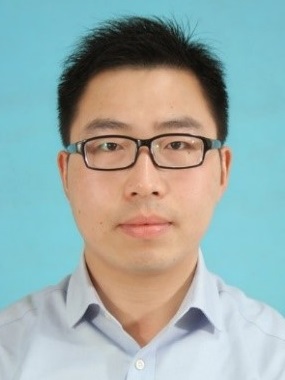}}]{Xun Yang} received the Ph.D. degree from the School of Computer and Information, Hefei University of Technology, China, in 2018. He is currently a postdoctoral research fellow with the NExT++ Research Center, National University of Singapore, Singapore. His current research interests include information retrieval, multimedia content analysis, and computer vision. He has served as the PC member and the invited reviewer for top-tier conferences and prestigious journals including ACM MM, IJCAI, AAAI, the ACM Transactions on Multimedia Computing, Communications, and Applications, IEEE Transactions on Neural Networks and Learning Systems, IEEE Transactions on Knowledge and Data Engineering, and IEEE Transactions on Circuits and Systems for Video Technology.
\end{IEEEbiography}

\begin{IEEEbiography}
[{\includegraphics[width=1in,height=1.25in,clip,keepaspectratio]{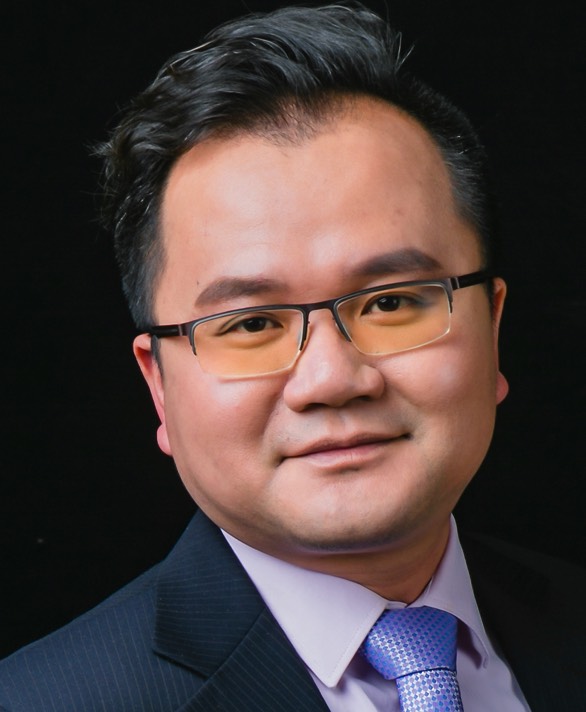}}]{Shuhui Wang}  received the B.S. degree in electronics engineering from Tsinghua University, Beijing, China, in 2006, and the Ph.D. degree from the Institute of Computing Technology, Chinese Academy of Sciences, Beijing, China, in 2012. He is currently a Full Professor with the Institute of Computing Technology, Chinese Academy of Sciences. He is also with the Key Laboratory of Intelligent Information Processing, Chinese Academy of Sciences. His research interests include image/video understanding/retrieval, cross-media analysis and visual-textual knowledge extraction.
\end{IEEEbiography}

\begin{IEEEbiography}[{\includegraphics[width=1in,height=1.25in,clip,keepaspectratio]{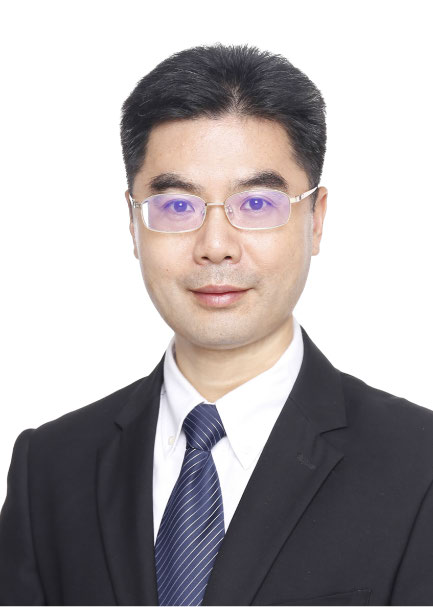}}]{Xun Wang} (Member, IEEE) received the B.S. degree in mechanics and the Ph.D. degrees in computer science from Zhejiang University, Hangzhou, China, in 1990 and 2006, respectively. He is currently a professor at the School of Computer Science and Information Engineering, Zhejiang Gongshang University, China. His current research interests include mobile graphics computing, image/video processing, pattern recognition, intelligent information processing and visualization. He is also a member of the ACM, and a senior member of the CCF.
\end{IEEEbiography}


\begin{IEEEbiography}[{\includegraphics[width=1in,height=1.25in,clip,keepaspectratio]{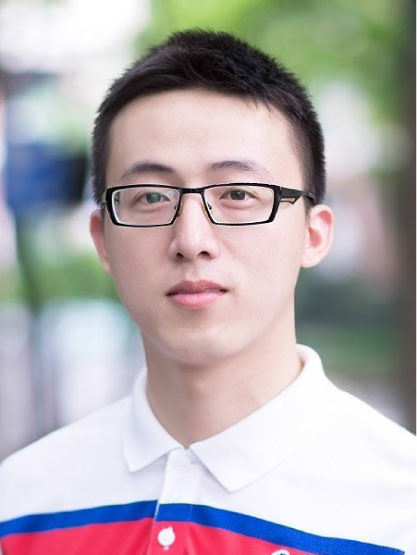}}]{Jianfeng Dong}
received the B.E. degree in software engineering from the Zhejiang University of Technology, China, in 2013, and the Ph.D. degree in computer science from Zhejiang University, China, in 2018. He is currently a Research Professor with the College of Computer Science and Technology, Zhejiang Gongshang University, Hangzhou, China. His research interests include multimedia understanding, retrieval, and recommendation. He was awarded the ACM Multimedia Grand Challenge Award and was selected into the Young Elite Scientists Sponsorship Program by the China Association for Science and Technology.
\end{IEEEbiography}

\end{document}

%% file: intro.tex
Traditional text-to-image retrieval \cite{wu2018learning,messina2021fine,pan2023fine,jin2024end,zhou2025achieving,liu2023efficient,zha2025ucpm,zhou2025dual,yang2024rebalanced} aims to find the most relevant image according to a single given textual description, having been a long-standing task in vision–language research.
However, in real-world scenarios, a single query is often insufficient to express the user’s full intent, where users typically refine their descriptions interactively, rather than articulating all details at once. To better capture the evolving user intent, chat-based image retrieval~\cite{levy2023chatting} has emerged as a natural extension of traditional text-to-image retrieval. Rather than executing a single-step search, this task engages in an interactive dialogue with the user, where it asks follow-up questions and incorporates the user’s answers to progressively narrow down the search space as illustrated in Fig.~\ref{fig:task}. 

\begin{figure*}[tb!]
\centering\includegraphics[width=1.98\columnwidth]{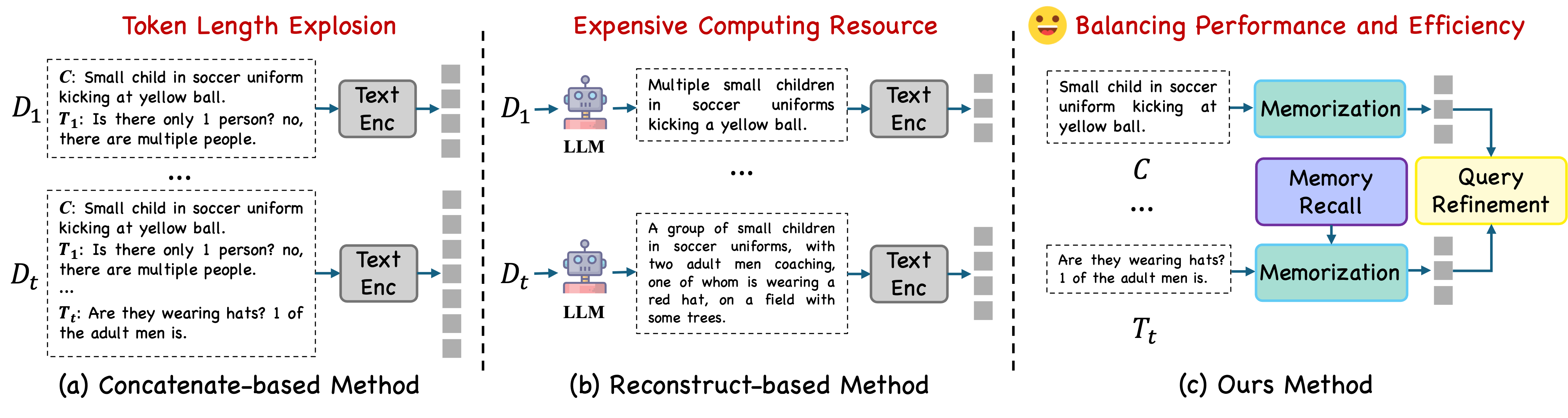}
\caption{Illustration of our motivation. Unlike previous methods that simply integrate dialogue contexts via either multi-round concatenation or LLM-based reconstruction, we propose an efficient memory-based approach that  enables more coherent and accurate understanding of evolving query intent.}\label{fig:intro}
\end{figure*}

Multiple studies~\cite{levy2023chatting,lee2024interactive,zhu2024enhancing,long2025diffusion,bai2025chat,chen2025mai} have demonstrated that this multi-round interaction paradigm can significantly enhance retrieval accuracy by iteratively refining the query representation.
While this interactive paradigm offers greater flexibility than single-round retrieval, it also poses new challenges, as the query intent continuously evolves throughout the dialogue, requiring the model to accurately and dynamically integrate and retain information across rounds.
Despite the advances, existing chat-based retrieval frameworks~\cite{levy2023chatting,lee2024interactive,long2025diffusion,luo2025imagescope} remain limited by their inefficient design of query intent understanding across dialogue rounds. As illustrated in Figure \ref{fig:intro} (a) and (b), current methods mainly follow two types of paradigms: the concatenate-based method~\cite{levy2023chatting} directly concatenates the initial query with all previous dialogue rounds into a single long text sequence to produce the query intent representation. This strategy leads to redundant computation over overlapping content, and as the concatenated text grows longer, both the difficulty of intent understanding and the computational burden increase rapidly.  In contrast, the reconstruct-based method \cite{lee2024interactive,luo2025imagescope,long2025diffusion} employs large language models (LLMs)~\cite{achiam2023gpt,ouyang2022training,touvron2023llama} to reconstruct the accumulated dialogue into a concise query at each round. Although this design can better capture the evolving intent than simple concatenation, it introduces heavy generative overhead and considerable additional computation from LLMs, thereby limiting overall efficiency.

To alleviate the above limitations, in this paper, we propose a Memory-Augmented Query Intent Understanding (MAQIU) framework that maintains a persistent and lightweight semantic memory to accumulate and refine query intent across dialogue rounds, and complements it with a memory recall mechanism to counter forgetting, thereby enabling efficient and effective interactive multi-round retrieval.
Instead of repeatedly encoding or reconstructing the entire dialogue, MAQIU performs progressive intent modeling, allowing each new query to interact with the retained memory to integrate new semantics while preserving accumulated intent.
However, regular long-term accumulation may still cause semantic drift or partial forgetting as the dialogue extends.
To address this, MAQIU stores all dialogue-round representations as historical context and selectively recalls informative past semantics to reinforce the current memory state, recovering fading intents and maintaining a balanced representation between newly integrated and long-term knowledge.
Additionally, existing methods often treat each round as an independent retrieval step, overlooking the feedback and continuity between rounds.
To enhance cross-round continuity, MAQIU further introduces a visual feedback mechanism that leverages image retrieval results from previous rounds to guide the formation of subsequent query representations.
By integrating retrieval-guided visual cues with our memory mechanisms, MAQIU achieves coherent and precise intent modeling across rounds while maintaining nearly constant computational complexity.

In summary, our contributions are as follows:
\begin{itemize}
    \item We propose a novel MAQIU framework for chat-based image retrieval, which introduces a memorization mechanism to progressively accumulate and refine query intent across dialogue rounds, together with a memory recall mechanism to mitigate semantic forgetting, enabling efficient and effective query intent understanding.
    \item We further incorporate a visual feedback mechanism that adopts image retrieval results from previous rounds to enhance cross-round continuity.
    \item We conduct extensive experiments on multiple dialogue-based benchmarks. MAQIU achieves a top-10 retrieval success rate of 83.21\% after 5 dialogue rounds and 87.45\% after 10 rounds, surpassing existing chat-based retrieval methods by a clear margin,while reducing dialogue encoding FLOPs by 86.4\% compared with the concatenate-based baseline ChatIR\cite{levy2023chatting}.
\end{itemize}

%% file: rel_work.tex
\subsection{Chat-based Image Retrieval}
To overcome the limitation of conventional text-to-image retrieval~\cite{yang2017pairwise,zhang2020context,messina2021fine,li2021align,jia2021scaling,diao2024gssf,liang2024multi,yang2023composed}, chat-based image retrieval~\cite{levy2023chatting} reformulates the task into an interactive paradigm that engages in multi-round dialogue.
By asking clarifying questions and integrating user responses, the system progressively obtains a refined query intent, which is then used to narrow the search space.
A key challenge of it is accurately modeling and integrating multi-round query intent. ChatIR~\cite{levy2023chatting} simply concatenate all dialogue rounds for joint encoding, which accumulates context but quickly leads to redundancy and inefficiency as the dialogue grows.
To mitigate this, PlugIR~\cite{lee2024interactive} employs LLM-based reformulation to generate concise queries for each round, enhancing coherence but introducing considerable generative overhead.
Similarly, ImageScope~\cite{luo2025imagescope} also adopts semantic reconstruction, yet designs a more sophisticated semantic decomposition mechanism with LLMs, at the expense of increased computational cost.
Overall, recent studies explored explored multi-round query intent understanding via concatenation or LLM-based reconstruction, but they often incur redundant encoding or excessive computational cost, highlighting the need for a more effective chat-based image retrieval framework.

\vspace{-1mm}
\subsection{Memory Modeling}
In natural language and multimodal understanding, memory-based architectures~\cite{dai2019transformer,lewis2020retrieval,wang2023augmenting,yu2024prompting,he2025hmt} have been widely explored to enhance context retention~\cite{he2026fine,li2025mitigating,guo2025prompt} and reduce information forgetting~\cite{zhou2026star}.
Current approaches can be broadly categorized into external retrieval memory and recurrent or persistent memory mechanisms.
The former explicitly stores explicitly cache historical representations and selectively retrieve them when relevant~\cite{wang2023augmenting,borgeaud2022improving,izacard2023atlas}.
By combining parametric and non-parametric storage, these methods enable efficient access to long-range context.
In contrast, recurrent or persistent memory frameworks update latent representations over time instead of re-encoding complete histories~\cite{he2025hmt, dai2019transformer,rae2019compressive}.
Such designs propagate compressed hidden states across segments, achieving temporal continuity with nearly constant computational cost.
These advances demonstrate the effectiveness of memory mechanisms for long-term dependency modeling, motivating our design of a compact and persistent memory representation that evolves continuously with dialogue progression.

%% file: method.tex
\begin{figure*}[tb!]
\centering\includegraphics[width=1.98\columnwidth]{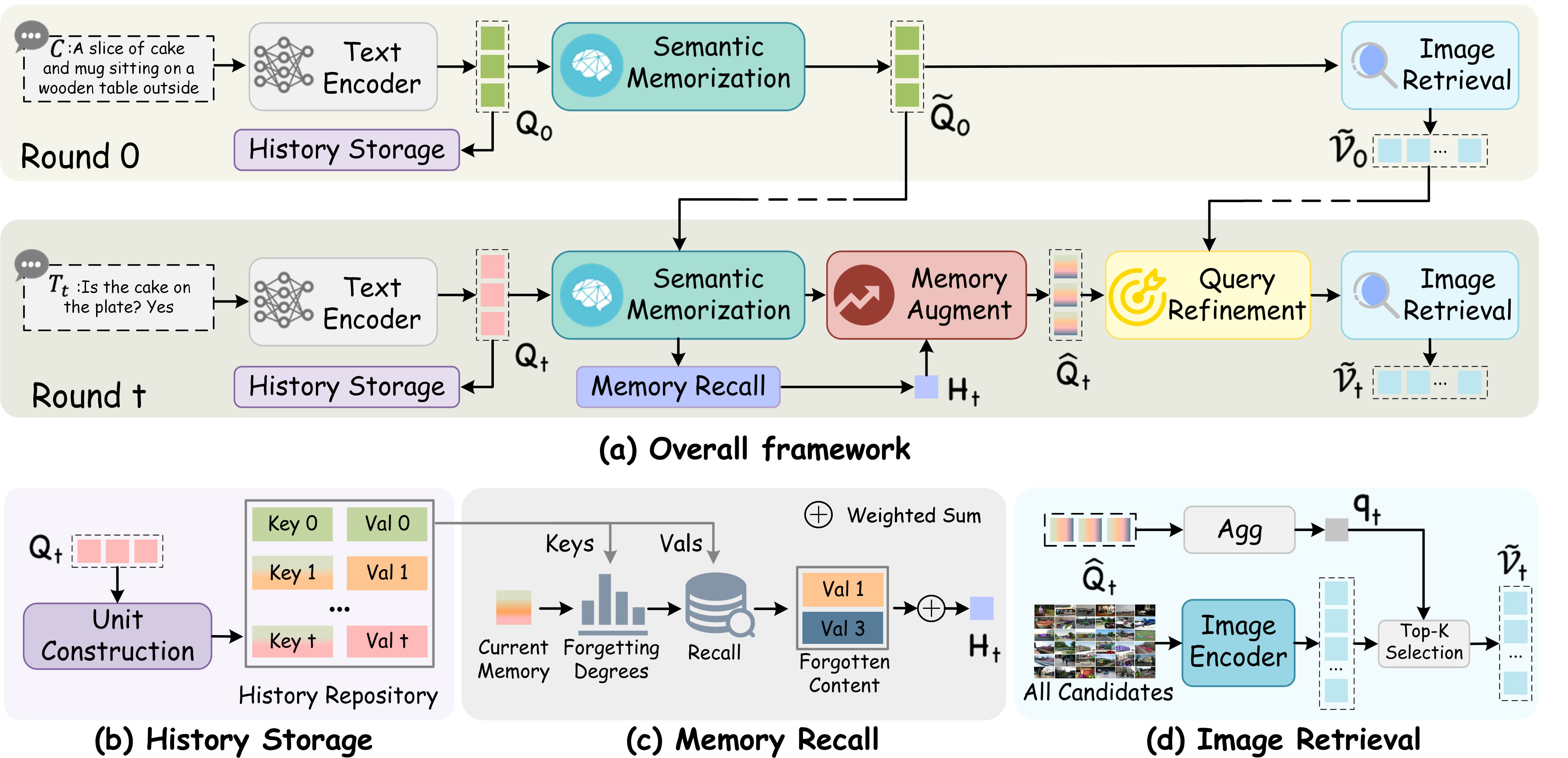}
\caption{Overview of the proposed MAQIU framework for chat-based image retrieval. Given the initial query and multi-round dialogue interactions, the model progressively updates a compact semantic memory to capture the evolving query intent without repeatedly encoding the full dialogue history. A history repository stores historical dialogue representations, while a memory recall module retrieves weakened semantics to reinforce the current memory state. Meanwhile, previous retrieval results are utilized as visual feedback to refine the current query representation and enhance cross-round semantic continuity. Finally, the refined query representation is used to retrieve relevant images, and the retrieved top-k results are propagated as visual history for subsequent dialogue rounds.}\label{fig:model}
\end{figure*}

\subsection{Preliminaries}
\noindent\textbf{Problem Definition.} The task of chat-based image retrieval aims to iteratively retrieve a target image from a large-scale image corpus through multi-round dialogue between the user and the retrieval system. Different from conventional text-to-image retrieval, where the user provides a single static textual description, chat-based image retrieval assumes that the user intent is gradually clarified and enriched during the dialogue process. Therefore, the retrieval model is required not only to match images with the current textual input, but also to continuously maintain and update a coherent representation of the user’s evolving search intent. 
Formally, let the image corpus be denoted as $\mathcal{I}=\{I_1,I_2,...,I_N\}$, where $N$ is the number of candidate images, and let $I_g \in \mathcal{I}$ be the target image that the user intends to retrieve. A dialogue session begins with an initial textual query $C$,  which usually provides a coarse description of the target image.
After that, the system interacts with the user for multiple rounds. At each round $t$, the system asks a clarifying question and receives a user response. We denote the textual content generated at the $t$-th round as $T_t$, which contains newly introduced semantic clues for refining the target image. After $t$ dialogue rounds, the accumulated dialogue context can be represented as $D_t=\{C,T_1,T_2,...,T_t\}$. Given $D_t$, the retrieval system computes a similarity score $s(I_i,D_t)$ between each candidate image $I_i \in \mathcal{I}$, and the current dialogue context, and then ranks all candidate images according to their relevance scores.


\noindent \textbf{Our Motivation.}
Existing chat-based image retrieval frameworks either repeatedly encode long dialogue histories or depend on large language models to reconstruct queries, leading to ineffective user intent understanding.
To address the challenge of continuously modeling and preserving the evolving query intent across dialogue rounds, we design a memory-augmented framework with three core components as illustrated in Figure \ref{fig:model}: (1) Progressive Dialogue-Semantic Memorization. The framework dynamically updates the memory with search semantics from the new dialogue round, forming a continuous intent representation without redundant re-encoding.
(2) Recalling Weakened Dialogue Memories. It maintains long-term semantic stability by selectively recalling informative historical contexts to counter forgetting and mitigate semantic drift.
(3) Query Refinement with Visual History. It refines the current query representation by incorporating visual feedback from previous image retrieval results, thereby enhancing cross-round coherence and maintaining semantic continuity throughout the dialogue. 
We will illustrate the details of each component in the following.

\subsection{Progressive Dialogue-Semantic Memorization}
In chat-based image retrieval, the user’s intent is not expressed in a single query but gradually unfolds as the round progresses. Therefore, an effective retrieval model should be able to continuously integrate newly provided information while preserving the intent semantics accumulated from previous rounds. Directly concatenating all historical dialogues leads to repeated encoding of redundant content, while independently encoding each round fails to maintain a coherent long-term intent representation. To address this issue, we introduce a progressive dialogue-semantic memorization mechanism, which represents the evolving query intent with a fixed set of memory tokens and updates them round by round.


Formally, given the textual query $C$ at the beginning of the dialogue, we encode it using a text encoder to obtain word-level textual features $Q_0\in \mathbb{R}^{n_q \times d_q}$, where $n_q$ denotes the number of textual words and $d_q$ is the feature dimension. To construct an initial memory state, we introduce a set of learnable memory tokens $E=\{e_i\}_{i=1}^{l}\in \mathbb{R}^{l \times d_q}$, where $l$ is the number of memory tokens. We then employ a transformer-based memorization module~\cite{vaswani2017attention} to inject the semantics of the initial query into the memory tokens. 
Specifically, the memory tokens $E$ are used as queries, while the initial textual features $Q_0$ are used as keys and values in a cross-attention operation:
\begin{equation}
    \widetilde{Q}_0=X \text{-}Attn(E,Q_0,Q_0),
\end{equation}
where $X\text{-} Attn(Q,K,V)$ denotes a standard cross-attention operation. Through this interaction, we form the initial dialogue memory state $\widetilde{Q}_0 \in \mathbb{R}^{l \times d_q}$ for subsequent multi-round information updates.


As the dialogue proceeds, the memory state is progressively updated with the newly introduced semantics from each dialogue round. For the $t$-th round ($t\ge 1$), the round text $T_t$ is encoded by the same text encoder to obtain textual features $Q_t\in \mathbb{R}^{n_r \times d_q}$, where $n_r$ denotes the word length of the current round text. 
Instead of re-encoding the entire dialogue history, we only encode the current round and use it to update the memory propagated from the previous round. Specifically, the augmented memory tokens $\widehat Q_{t-1}$, which are obtained after the memory recall process (Introduced in Sec.~\ref{sec:mem_recall}), interact with the current textual features through cross-attention:
\begin{equation}
    \widetilde{Q}_t=X \text{-} Attn(\widehat{Q}_{t-1},Q_t,Q_t).
\end{equation}
Through this operation, the memory tokens absorb the newly introduced semantics while preserving previously accumulated intent knowledge. Since the number of tokens $l$ is fixed, this iterative update allows our model to maintain constant computational complexity across dialogue rounds while progressively refining the representation of user intent.

\subsection{Recalling Weakened Dialogue Memories }\label{sec:mem_recall}
Although the progressive memorization mechanism can efficiently update the query intent representation across dialogue rounds, the iterative update process may still suffer from semantic weakening. As new dialogue information is continuously injected into the memory tokens, recently introduced semantics may dominate the memory state, while some useful clues from earlier rounds are gradually diluted or overwritten.
To preserve the completeness of the evolving query intent, we introduce a memory recall mechanism that stores historical dialogue representations and selectively recalls potentially weakened semantics to augment the current memory state.

\subsubsection{Historical Dialogue Repository Construction} 
To enable long-term semantic recall, we construct a historical dialogue repository that preserves information from all previous dialogue rounds. Each historical unit is organized in a $(key:val)$ format, where the key is used to estimate the relevance between a historical dialogue and the current memory state, and the value stores the corresponding dialogue representation for later augment. This design decouples the selection of historical dialogues from the retrieval of their semantic content, enabling more flexible and reliable memory recall.
For the dialogue representation $Q_t$ at round $t$, we do not directly use its global textual embedding as the key. Instead, we first enhance its interaction with the initial memory state $\widetilde{Q}_0$, since the key is later compared with the evolving memory state for forgetting estimation. This memory-aware interaction makes the key more compatible with memory representations and thus improves the reliability of historical recall. Specifically, we perform cross-attention between $\widetilde{Q}_0$ and $Q_t$:
\begin{equation}
    M_t = X \text{-} Attn(\widetilde{Q}_0,Q_t,Q_t).
\end{equation}
The output $M_t$ is further aggregated through an attention pooling to produce a global vector ${m}_{t}^* \in \mathbb{R}^{1 \times d_q}$ as the $key$:
\begin{equation}\label{eq:attn}
    \begin{aligned}
    m_t^* =\sum_{i=1}^{l} {\alpha_{i}}\times{m_{i}}, \quad
    \alpha=Softmax({M_t} w),
    \end{aligned}
\end{equation}
where $Softmax$ denotes the softmax layer, $w\in\mathbb{R}^{d_q\times{1}}$ is the trainable vector, and $\alpha \in\mathbb{R}^{{1 \times l} } $ indicates the attention vector.
Then we take $Q_t^{cls}$ as corresponding $val$, which represents the global textual embedding derived from the $cls$ token of the text encoder, capturing the semantics of the current round.
The $({m}_{t}^*:Q_t^{cls})$ pair is then collected to form the historical dialogue repository for subsequent recall as:
\begin{equation}
\mathcal{M} = \{(m_0^*:Q_0^{cls}),({m}_{1}^*:Q_1^{cls}), \dots, ({m}_{t}^*:Q_t^{cls})\}.
\end{equation}

\subsubsection{Forgotten Content Recall and Augment}
As dialogue rounds progress, faraway historical dialogue semantics may gradually fade from the current memory due to long-time updates.
To mitigate long-term forgetting, we retrieve these semantically distant historical representations from the repository and reintegrate them into the current memory to achieve semantic completeness.
Specifically, we first aggregate the memory tokens of current round $\widetilde{Q}_t$ through attention pooling same as Eq.(\ref{eq:attn}) to obtain $\widetilde{q}_t^{*} \in \mathbb{R}^{1 \times d_q}$ as the global representation of the current memory state. 
Historical dialogues with lower similarity to the current memory are more likely to contain forgotten or overwritten information.
To quantify the forgetting degree of each historical dialogue and preserve the semantic completeness of the query intent, we compute the cosine similarity between $\widetilde{q}_t^{*}$ and all stored keys $\{m_i^*\}_{i=1}^{t-1}$ in the repository $\mathcal{M}$, and select the $n$ least similar ones as complementary references:
\begin{equation}
\mathcal{J} = \text{Min-}n\big(\{\text{cos}(\widetilde{q}_t^{*}, m_i^*)\}_{i=1}^{t-1}\big),
\end{equation}
where $\text{cos}$ denotes cosine similarity. For the selected keys, we compute normalized recall weights based on the inverse similarity:
\begin{equation}
w_i = 1- \frac{\text{exp}(\text{cos}(\widetilde{q}_t^{*}, m_i^*))}{\sum_{j \in \mathcal{J}} (\text{exp}(\text{cos}(\widetilde{q}_t^{*}, m_j^*)))}, \quad i \in \mathcal{J}.
\end{equation}
The corresponding values $\{Q_i^{cls}\}_{i \in \mathcal{J}}$ are aggregated into a weighted recalled representation $H_t \in \mathbb{R}^{1 \times d_q}$ and then used to augment the current memory state via cross-attention, as:
\begin{equation}
\begin{aligned}
    \widehat{Q}_t=X \text{-} Attn(\widetilde{Q}_{t},H_t,H_t), \quad
H_t = \sum_{i \in \mathcal{J}} w_i Q_i^{cls}.
\end{aligned}
\end{equation}
Through this recall-augment process, the current memory state can recover complementary information from earlier dialogue rounds while retaining the newly updated semantics. As a result, the model alleviates semantic forgetting and maintains a more complete representation of the user’s long-term query intent across multi-round interactions. This augmented memory $\widehat{Q}_t$ is then passed to the subsequent retrieval and query refinement stages.

\subsection{Query Refinement with Visual History}
Beyond textual dialogue history, previous retrieval results also provide useful visual cues for understanding the evolving query intent. In chat-based image retrieval, consecutive dialogue rounds are naturally correlated, as they are all conducted to refine the retrieval of the same target image. Therefore, the top-ranked images from earlier rounds can be viewed as coarse visual feedback, reflecting the visual semantics emphasized by the model at that stage. Since our image and text encoders are initialized from the strong vision-language model BLIP~\cite{li2022blip}, early-round retrieval results often contain candidate images with semantics relevant to the user query. Thus, previous top-ranked results are not simply random noise, but can provide visual patterns complementary to the textual dialogue context. Motivated by this observation, we introduce a visual history refinement mechanism that incorporates previous retrieval results into the current query representation to enhance cross-round semantic continuity.
Specifically, after completing the retrieval process of the previous round $t{-}1$, we record the corresponding retrieval feedback as visual history set $\mathcal{V}_{t-1}$. Each candidate image $I_j$ in the image corpus $\mathcal{I}$ is encoded by the image encoder to obtain its visual embedding $v_j \in \mathbb{R}^{d}$.
We then compute cosine similarities between the final query representation of the previous round $q_{t-1}$ and all image embeddings $\{v_j\}_{j=1}^{N}$, and select the top-$k$ images with the highest similarity to form  $\mathcal{V}_{t-1} \in \mathbb{R}^{k \times d}$ 
as:
\begin{equation}
\mathcal{V}_{t-1} = \text{Top-}k\big({ \{\text{cos}(q_{t-1}, v_j) }\}_{j=1}^{N} \big).
\end{equation}
$\mathcal{V}_{t-1}$ is then projected into the same feature space as the current memory tokens $\widehat{Q}_t$ via a fully connected layer (FC), and fused through a transformer-based cross-attention module:
\begin{equation}
    \begin{aligned}
        \bar{Q}_t=X \text{-} Attn(\widehat{Q}_{t},\widetilde{\mathcal{V}}_{t-1},\widetilde{\mathcal{V}}_{t-1}), \quad
    \widetilde{\mathcal{V}}_{t-1} = FC(\mathcal{V}_{t-1}).
    \end{aligned}
\end{equation}
The fused representation $\bar{Q}_t$ is further aggregated by attention pooling (same as Eq.(\ref{eq:attn})) into a global query vector and projected to the same embedding dimension as the image features, yielding the final query representation $q_t \in \mathbb{R}^{d}$ of current round $t$. Finally, $q_t$ is used to compute similarity scores with all image embeddings, retrieving the current top-$k$ results that provide visual feedback for the next dialogue round.
This visual-historical refinement establishes a cross-round feedback loop, effectively strengthening the semantic connection between dialogue rounds.

\subsection{Training Objective}

We train the model end-to-end using a contrastive loss~\cite{faghri2017vse++, radford2021learning} that aligns dialogue-aware textual representations with their corresponding visual embeddings.
At each dialogue round, the model computes a retrieval loss that encourages the current query representation to be closer to its matched image while pushing it away from mismatched ones within the same batch.

Formally, given the final query representation $q_t$ at round $t$ and the visual embeddings $\{v_j\}_{j=1}^{B}$ of all images in a mini-batch of size $B$, the loss for this round is defined as:
\begin{equation}
\begin{aligned}
    \mathcal{L}_t
= -\frac{1}{2B}
\sum_{i=1}^{B}
\Big[
\log \frac{\exp(\text{cos}(q_i,v_i)/\tau)}{\sum_{j=1}^{B} \exp(\text{cos}(q_i,v_j)/\tau)}
+\\
\log \frac{\exp(\text{cos}(v_i,q_i)/\tau)}{\sum_{j=1}^{B} \exp(\text{cos}(v_i,q_j)/\tau)}
\Big],
\end{aligned}
\end{equation}
where $\tau$ is a temperature parameter.
This contrastive objective is applied independently at each dialogue round, driving the model to align textual and visual features progressively as the dialogue evolves.
The final training objective averages the losses over all dialogue rounds:
\begin{equation}
\mathcal{L} = \frac{1}{R} \sum_{i=1}^{R} \mathcal{L}_{i},
\end{equation}
where $R$ is the total number of dialogue rounds.
This round-averaged optimization ensures that the model learns to maintain consistent retrieval performance throughout the multi-turn interaction rather than overfitting to any single round.

%% file: experiment.tex
\section{Experiments}

\subsection{Experimental Setup}
\noindent\textbf{Datasets.} Following previous works~\cite{levy2023chatting,lee2024interactive}, we conduct our main experiments under the chat-based image retrieval setting using the human-annotated dialogue-based benchmark Visual Dialog (VisDial)~\cite{das2017visual,murahari2020large}. 
Each image in VisDial is paired with an initial query and ten rounds of dialogues, and the dataset contains 123,287 images for training and 2,064 images for validation.
Beyond the official VisDial benchmark, we further evaluate our method on two cross-domain dialogue sets introduced in~\cite{levy2023chatting}, namely ChatGPT-BLIP2 and Human-BLIP2.
Both datasets are constructed upon the validation images of VisDial but contain newly generated dialogues derived from question–answer generation sources.

\noindent\textbf{Evaluation metrics.} Previous works~\cite{levy2023chatting,lee2024interactive} mainly adopt Hit@10 as the evaluation metric,
which measures the proportion of samples whose target image appears within the top-10 retrieved results at any dialogue round.
However, Hit@10 alone cannot faithfully capture the instantaneous retrieval accuracy at each round. Since it accumulates success over all previous rounds, it is a non-decreasing metric by design, once a target is retrieved within the top-10, it is considered a successful case for all subsequent rounds, regardless of later ranking degradation. Consequently, a model that frequently changes retrieval rankings or occasionally produces large variations may achieve a high Hit@10 even when its per-round retrieval quality is inconsistent.
To provide a more precise evaluation of round-wise retrieval quality, we also utilize Recall@10, which measures the proportion of targets ranked within the top-10 in the current round only.
Moreover, to balance between cumulative success and instantaneous accuracy, we report the mean of Hit@10 and Recall@10 at each round. For ease of reference, we denote it as MHR@10. We also report the average performance of ten rounds as the whole evaluation.

\noindent \textbf{Implementation Details.} Our framework is implemented in PyTorch and trained on 2 A100 GPUs.
Both the text and image encoders in our model are initialized from BLIP~\cite{li2022blip}.
The number of memory tokens $l$ is fixed at 36, the recall number n is set to 2 and the number of visual history $k$ is set to 100.
Additionally, the memory recall mechanism is activated only when the current round index $t \ge 3$. 
The dimensionality of the text encoder output $d_q$ is 768, while the image embeddings and the final projected representations $d$ are 256.
For model training, we utilize an AdamW optimizer with a mini-batch size of 256.
We optimize both the backbone and the task-specific head of model.
The backbone involves only the text encoder, trained with a learning rate 1.25e-5, while the head modules  are trained with a learning rate 1e-4. The maximal number of epochs is set to 36.

\begin{table*}[tb!]
\renewcommand{\arraystretch}{1.2}
\caption{Performance comparison on VisDial under two dialogue sources: the human-annotated original dialogue and the LLM-generated reconstructed dialogue. We report MHR@10 in the following table and the best results are in bold.}
\centering
\label{tab:visdial perform}
\scalebox{1.1}{
\begin{tabular}{lccccccccccc}
\toprule
\multicolumn{1}{c}{\multirow{2}{*}{\textbf{Method}}} & \multicolumn{11}{c}{\textbf{\#Round (MHR@10)}} \\ 
\multicolumn{1}{c}{} & 1 & 2 & 3 & 4 & 5 & 6 & 7 & 8 & 9 & 10 & Avg \\ \hline
\textbf{Original dialogue:} &  &  &  &  &  &  &  & &  &  & \\
Zero-shot BLIP~\cite{li2022blip} &  72.90&  73.21&  72.94&  73.09& 73.28 & 73.26 & 72.63 & 72.55 & 72.39 & 72.19 & 72.84\\
ImageScope~\cite{luo2025imagescope} &68.36 & 71.95 & 73.31 & 75.51 & 75.93 & 76.75 & 77.38 & 78.01 & 77.23 & 78.49 & 75.29 \\
ChatIR~\cite{levy2023chatting}  & 73.30 & 76.67 & 79.12 & 80.67 & 81.64 & 82.82 & 83.58 & 84.11 & 84.50 & 85.51 &81.19 \\
MAQIU (ours)  & \textbf{75.39} & \textbf{78.15} & \textbf{80.62} & \textbf{81.91} & \textbf{83.21} & \textbf{84.42}  & \textbf{85.24} & \textbf{85.83} & \textbf{86.41} & \textbf{87.45} & \textbf{82.86} \\ 
\cline{1-12}
\textbf{Reconstructed dialogue:} &  &  &  &  &  &  &  & &  & &  \\
Zero-shot BLIP~\cite{li2022blip} &  74.56&  76.99&  78.10&  78.83& 78.93 & 78.85 & 78.03 & 77.30 & 76.89 & 76.48 & 77.50\\
ChatIR~\cite{levy2023chatting} & 74.47  &76.84  & 78.52 & 80.23 & 80.91 & 82.20 & 82.64 &82.44 & 83.09 &83.71 &  80.51\\
PlugIR~\cite{lee2024interactive}  & 76.65 & 79.29 & 81.03 & 81.83 & 82.75 & 82.78 & 83.10 & 82.90 &82.73  &82.92 & 81.60 \\
MAQIU (Ours) & \textbf{77.21} & \textbf{79.80} & \textbf{81.47} & \textbf{82.61} & \textbf{83.10} &\textbf{83.90} &\textbf{84.33} &\textbf{84.67} &\textbf{84.98} &\textbf{84.96} & \textbf{82.70}\\
\bottomrule
\end{tabular}
}
\end{table*}

\begin{figure*}[tb!]
\centering
\subfloat[Original dialogue]{
\includegraphics[width=0.9\columnwidth]{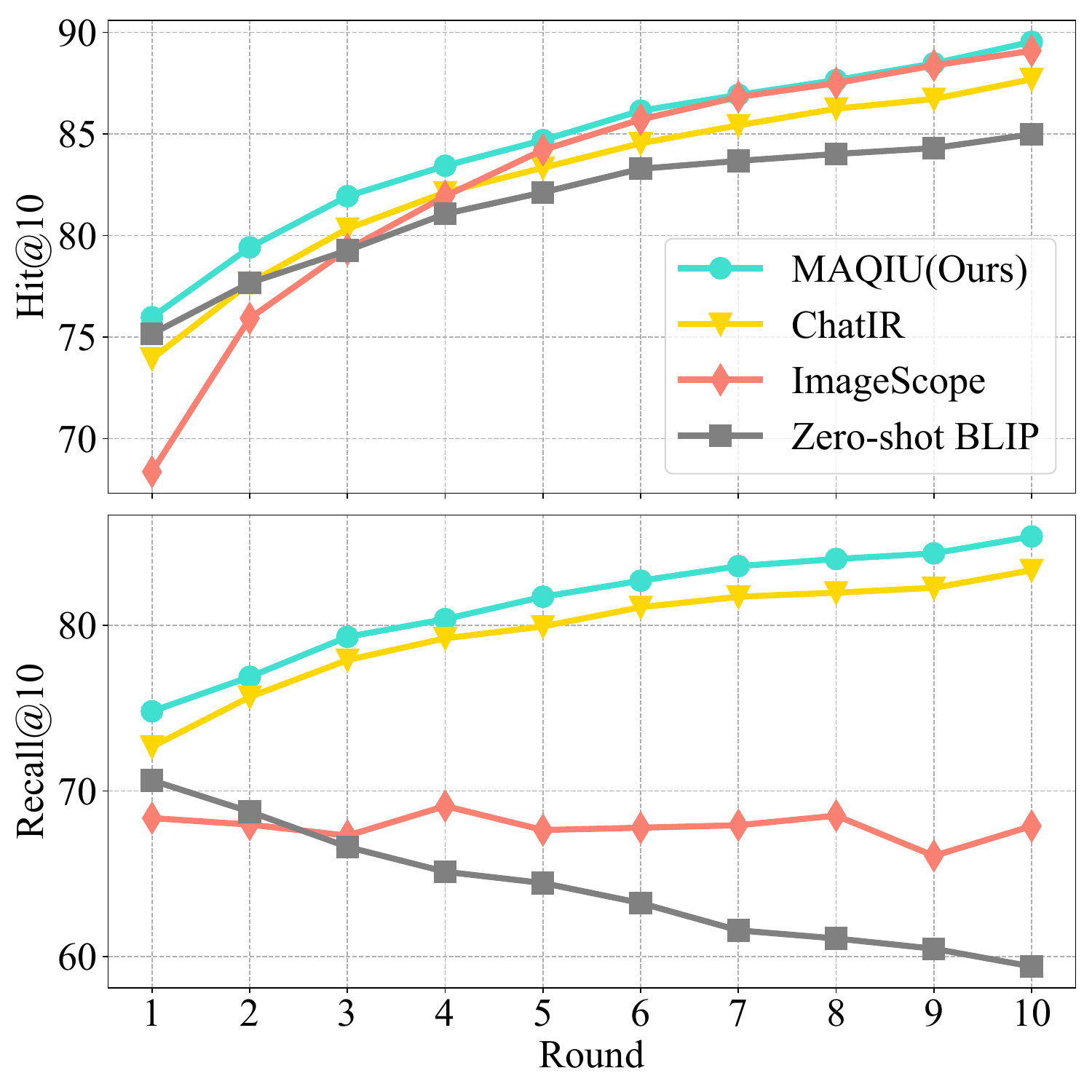}
}
\hspace{0.04\textwidth}
\subfloat[Reconstructed dialogue]{
\includegraphics[width=0.9\columnwidth]{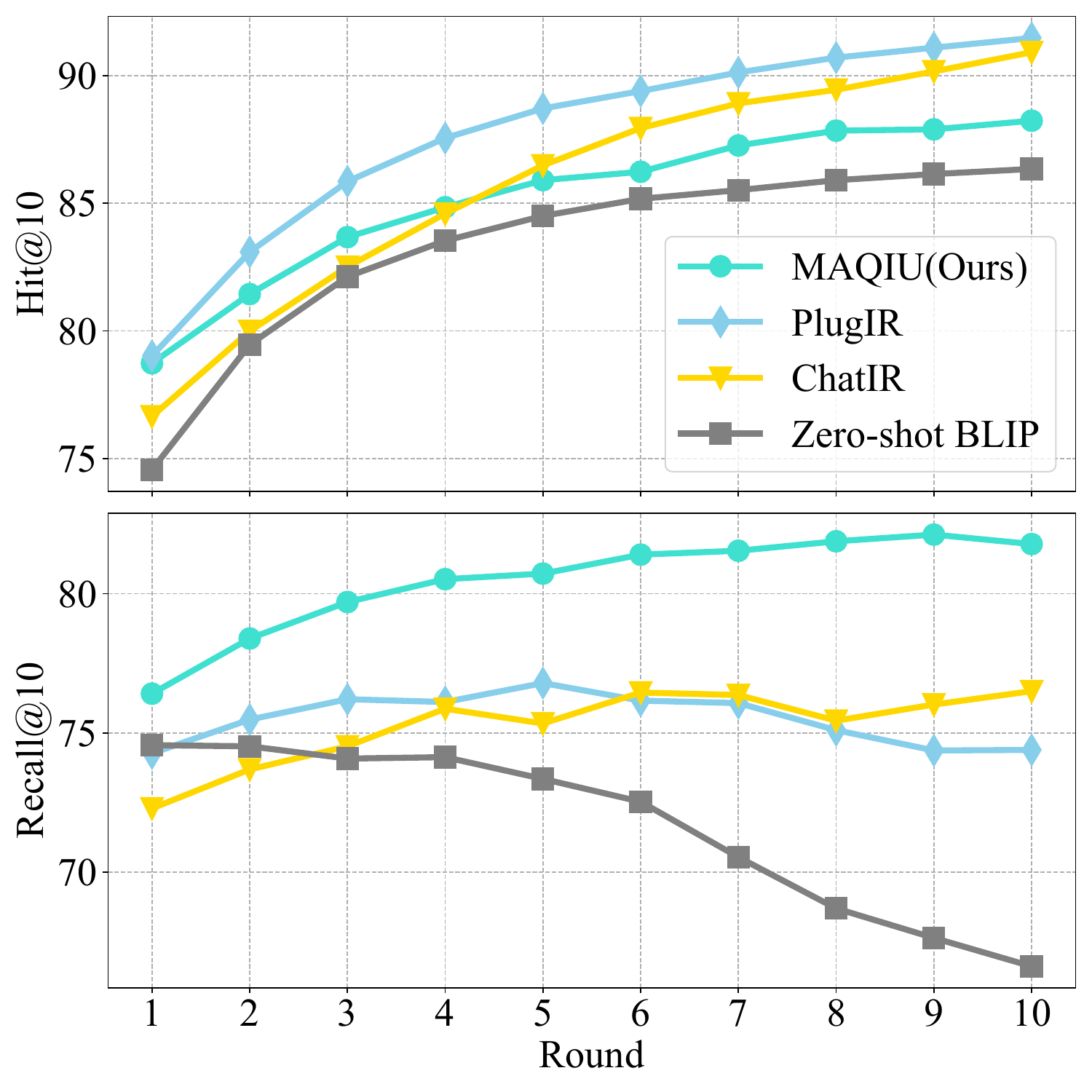}
}
\caption{Comparison of Hit@10 and Recall@10 across dialogue rounds on VisDial under Original (a) and Reconstructed (b) dialogue settings. While Hit@10 steadily improves for all methods, many baselines exhibit fluctuating or degrading Recall@10, whereas our method maintains consistently rising recall, demonstrating stronger long-term semantic retention.}\label{fig:hit and recall}
\vspace{-3mm}
\end{figure*}

\subsection{Performance Comparison}
We first compare the proposed MAQIU framework with representative state-of-the-art methods on the VisDial dataset.
In addition to evaluating on the original dialogues of VisDial, we also adopt the reconstructed dialogue released by \cite{lee2024interactive} for a more comprehensive performance assessment.
Table~\ref{tab:visdial perform} summarizes the MHR@10 scores across ten dialogue rounds.
Overall, MAQIU consistently achieves the highest accuracy in all rounds and dialogue types, demonstrating its strong capability for multi-round query understanding and reasoning. On the original dialogue, the best-performing method ChatIR concatenates the entire dialogue history with the current query, which yields early gains but quickly introduces redundant or conflicting context.
Unlike the concatenation-based design, MAQIU progressively integrates newly introduced query intents, and thus achieves more stable and sustained performance improvement throughout the dialogue. 
On the reconstructed Dialogue, PlugIR achieves competitive Hit@10 performance but exhibits limited growth under the MHR@10 metric, indicating a gradual decline in round-level retrieval accuracy.
This reveals a key limitation of Hit@10, whose cumulative nature tends to hide precision degradation in later rounds.
In contrast, MAQIU maintains a consistent upward trend across all rounds, showing its superior ability to continuously understand query intent and preserve retrieval precision as the dialogue evolves.

In addition, we provide detailed Hit@10 and Recall@10 results for all compared methods, as shown in Fig.~\ref{fig:hit and recall}. Across both dialogue settings, Hit@10 generally increases with dialogue rounds, indicating that multi-round interaction improves the likelihood of including the target image within the top-10 retrieved results.
However, on reconstructed dialogues, our Hit@10 is slightly lower than the baselines in some cases. This can be attributed to the fact that under reconstructed settings our memory-based encoding repeatedly incorporates reconstructed texts across dialogue rounds, which introduces redundant information and makes Hit@10 less indicative of long-term intent integration quality.
In contrast, the Recall@10 curves reveal more pronounced differences. While our method maintains steady improvement across rounds, several baselines show noticeable fluctuations or even clear declines in later rounds. These drops suggest that such methods struggle to preserve long-term intent information, often overemphasizing local dialogue updates and losing alignment with earlier semantic cues.

\begin{figure}[tb!]
    \centering
    \subfloat[ChatGPT-BLIP2]{
    \includegraphics[width=0.48\linewidth]{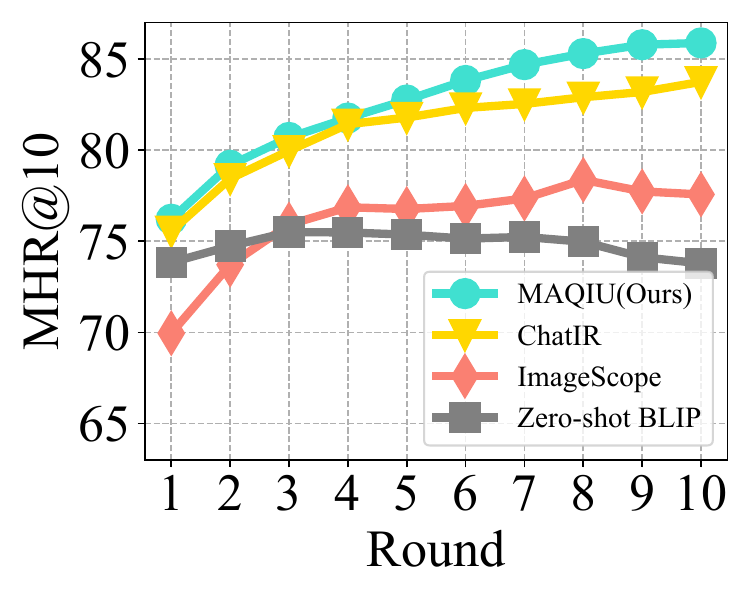}
    }
    \subfloat[Human-BLIP2]{
    \includegraphics[width=0.48\linewidth]{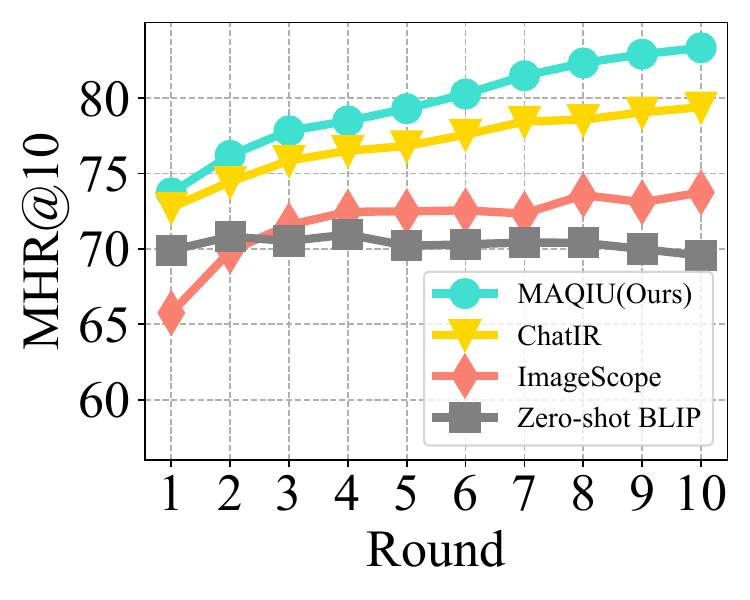}
    }
    \caption{Performance comparison on dialogue datasets ChatGPT-BLIP2 and Human-BLIP2. 
    The results demonstrate strong generalization of our proposed model in the cross-domain evaluation.}\label{fig:perform on other datasets}
\end{figure}

\subsection{Cross-domain Evaluation}
To further evaluate generalization beyond the VisDial domain, we conduct experiments on the ChatGPT-BLIP2 and Human-BLIP2 datasets.
Both datasets serve as cross-domain extensions of VisDial, containing dialogues whose linguistic form and semantic focus differ markedly from the original annotations.
As shown in Fig.~\ref{fig:perform on other datasets}, MAQIU achieves the highest MHR@10 on both datasets and maintains consistent gains across dialogue rounds.
The results demonstrate that MAQIU effectively generalizes across dialogue domains, exhibiting strong robustness to distributional variations in both human- and LLM-generated dialogues.


\subsection{Computational Complexity Analysis}
To assess the computational efficiency of MAQIU in chat-based image retrieval, we jointly analyze the growth of average token length and the corresponding FLOPs across dialogue rounds. The average token length is computed as the mean number of textual tokens encoded per round on the VisDial test set.
As shown in Table~\ref{tab:complexity}, the concatenate-based method continuously appends previous dialogues to the query, causing the token count to expand from 25 to 122 and the FLOPs to increase nearly fivefold by the tenth round.
The reconstruct-based method further amplifies this trend, as both the reconstruction prompt and LLM-generated text contribute to a sharp rise in token length and trillion-scale FLOPs, resulting in the heaviest computational load.
In contrast, MAQIU encodes only the current-round query while accumulating dialogue semantics through a fixed set of memory tokens, keeping both the token length and FLOPs nearly constant throughout the dialogue.
This confirms that MAQIU effectively suppresses computation growth and achieves stable efficiency for chat-based image retrieval. Beyond the FLOPs analysis, we also observe clear practical latency advantages at round 10. Specifically, for text-side processing, encoding the fully concatenated dialogue requires about 4 ms per query, while encoding the reconstructed text requires about 2 ms per query, in addition to the LLM-based reconstruction overhead of around 1–2 s per query in our implementation. In comparison, MAQIU requires only about 1 ms per query.

\begin{table*}[]
\renewcommand{\arraystretch}{1.2}
\centering
\caption{Computational complexity comparison in terms of average token
length and FLOPs. Our memory-based method exhibits substantially lower computational cost.}
\scalebox{1.0}{
\begin{tabular}{lcccccccccc}
\toprule
\multirow{2}{*}{} & \multicolumn{10}{c}{\textbf{\#Round}} \\
 & 1 & 2 & 3 & 4 & 5 & 6 & 7 & 8 & 9 & 10 \\ \hline
\multicolumn{11}{l}{\textbf{Average Tokens (Number) $\downarrow$ :}} \\
Reconstruct-based & 329 & 354 & 360 & 375 & 390 & 406 & 421 & 436 & 451 & 465 \\
Concatenate-based & 25 & 36 & 46 & 57 & 68 & 79 & 90 & 100 & 111 & 122 \\
Memory-based (Ours) & 11 & 11 & 10 & 11 & 11 & 11 & 11 & 12 & 12 & 12 \\ 
\hline
\multicolumn{11}{l}{\textbf{FLOPs (G) $\downarrow$:}} \\
Reconstruct-based & 115K & 120K & 126K & 131K & 137K & 142K & 147K & 153K & 158K & 163K \\
Concatenate-based & 4.3 & 6.2 & 7.9 & 9.8 & 11.7 & 13.6 &  15.6& 17.3 & 19.3 & 21.3 \\
Memory-based (Ours) & 2.9 & 2.9 & 2.9 & 2.9 & 2.9 & 2.9 & 2.9 & 2.9 & 2.9 & 2.9 \\
\bottomrule
\end{tabular}\label{tab:complexity}
}
\end{table*}

\subsection{Ablation Study}
\subsubsection{The effectiveness of proposed modules}
We evaluate the contribution of each component in MAQIU by starting from a baseline that includes only the Progressive Dialogue-Semantic Memorization (PDSM) module and progressively adding the Memory Recall (MR) and Query Refinement with visual history (QR) modules. The results are reported in Table~\ref{tab:ablation}.
Adding either MR or QR individually yields clear gains over the PDSM baseline, indicating their complementary effects in enhancing memory retention and cross-round semantic continuity.
Combining both MR and QR leads to the highest performance, confirming that the two modules are complementary and jointly enhance the retrieval performance.

\subsubsection{The effectiveness of progressive memorization}
To evaluate the isolated effectiveness of PDSM, we compare it with three alternative strategies for multi-round dialogue fusion:
(1) Similarity-based Aggregation (Sim-Agg), a non-iterative scheme that processes each round independently and aggregates all dialogue features via similarity weighting;
(2) Iterative Weighted Sum (IWS), which linearly combines the current and previous dialogue features;
(3) Iterative Concatenation Fusion (ICF), which concatenates current and previous dialogue features followed by an MLP projection.
As shown in Table ~\ref{tab:ablation pdsm}, PDSM consistently outperforms all variants.
Sim-Agg performs the worst, as its round-independent aggregation mixes heterogeneous dialogue signals and easily suppresses the dominant semantics derived from the initial query. IWS and ICF perform better, since they propagate information across rounds, but their global fusion operations remain coarse, making it difficult to preserve the fine-grained semantic evolution of the dialogue.
By contrast, PDSM preserves explicit memory tokens anchored to the initial query and updates them via cross-attention, enabling selective integration of new dialogue semantics while retaining accumulated intent and resulting in superior retrieval accuracy.

\begin{table}[]
\caption{Ablation studies of MAQIU. All components are effective, with their combination achieving the best performance.}
\renewcommand{\arraystretch}{1.2}
\centering
\scalebox{1.1}{
\begin{tabular}{ccccccc}
\toprule
\multicolumn{2}{c}{\textbf{Module}} &  & \multicolumn{4}{c}{\textbf{\#Round(MHR@10)}} \\ 
\cline{1-2} \cline{4-7} 
  MR & QR &  & 3 & 6 & 10 & Avg \\ 
 \cline{1-2} \cline{4-7} 
 \ding{55}  &\ding{55}   &  & 76.97 & 81.95 & 85.47 & 80.08 \\
\ding{51}  &\ding{55}   &  & 77.98 & 83.64 & 86.78  & 81.78 \\
 \ding{55}  &\ding{51}  &  & 79.21 & 83.01 & 86.37 & 81.38 \\
  \ding{51} &\ding{51}  &  & \textbf{80.62} & \textbf{84.42} & \textbf{87.45} & \textbf{82.86} \\
\bottomrule
\end{tabular}\label{tab:ablation}
}
\end{table}

\begin{table}[]
\renewcommand{\arraystretch}{1.2}
\centering
\caption{Performance comparison between PDSM and three dialogue semantic fusion baselines, our progressive memorization design outperforms all counterparts across rounds.}
\scalebox{1.1}{
\begin{tabular}{lccccc}
\toprule
\multirow{2}{*}{\textbf{Method}} &  & \multicolumn{4}{c}{\textbf{\#Round(MHR@10)}} \\ \cline{3-6} 
 &  & 3 & 6 & 10 & Avg \\ \cline{1-1} \cline{3-6} 
Sim-Agg & & 69.87  & 74.43 & 76.12 & 72.53 \\
ICF &  & 71.51 & 76.35 & 79.36 & 74.66 \\
IWS &  & 75.09 & 80.52 & 84.23 & 78.64 \\
Ours & & \textbf{76.97} & \textbf{81.95} & \textbf{85.47} & \textbf{80.08}   \\ 
\bottomrule
\end{tabular}\label{tab:ablation pdsm}
}
\end{table}

\begin{figure}[tb!]
\centering
\subfloat[]{\includegraphics[width=0.49\linewidth]{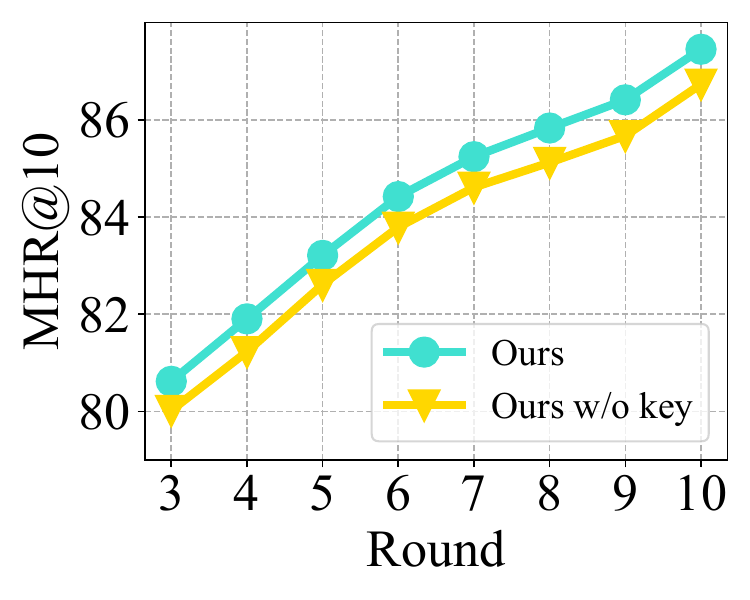}}
\subfloat[]{\includegraphics[width=0.49\linewidth]{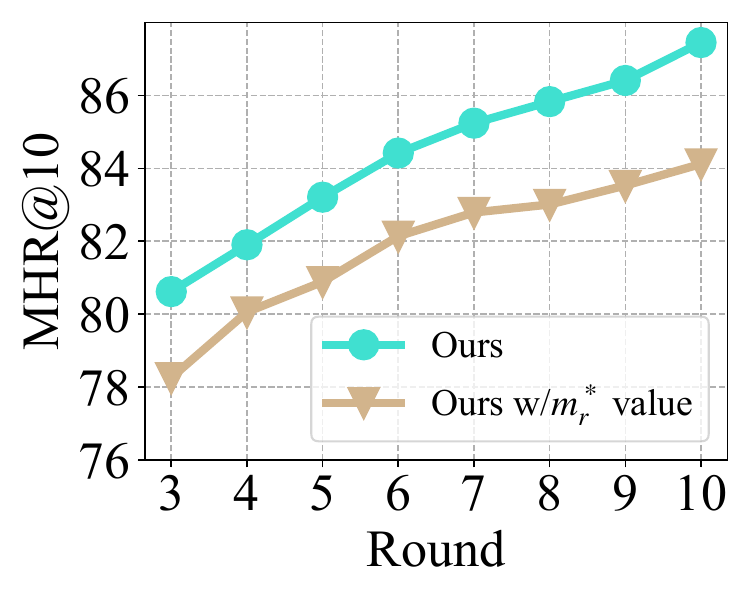}}
\caption{Ablation on historical dialogue repository construction in terms of (a) the key–value storage structure of historical dialogue and (b) the impact of value selection.}\label{fig:history construction}
\end{figure}

\subsubsection{The studies on historical dialogue repository}
We first study the effectiveness of dictionary structured dialogue storage. 
In our design, each historical dialogue is stored as a key–value pair $(m_t^*: Q_t^{cls})$, where $m_t^*$ is used for assessing the relevance of past dialogues to the current memory state, and $Q_t^{cls}$ preserves the dialogue representation used for potential recall. 
To validate the effectiveness of this structure, we compare it against a simplified alternative that removes the key entirely and stores $Q_t^{cls}$ only. As shown in 
Fig.~\ref{fig:history construction} (a), discarding the key leads to a consistent performance drop across dialogue rounds, indicating that relying solely on raw dialogue embeddings makes forgetting assessment less accurate. It also demonstrates that $m_t^*$ is more semantically aligned with memory states, enabling more reliable retrieval of forgotten information.

We further examine how the choice of stored values affects the recall process. While our default setting uses $Q_t^{cls}$ as the recall content, we replace it with $m_t^*$, as this representation is aligned with the semantic space of memory tokens. As shown in Fig.~\ref{fig:history construction} (b), this substitution results in clear performance degradation. The reason is that $m_t^*$ will repeatedly reintroduce the semantics of initial query into the memory tokens, weakening the model's ability to track semantic progression across the dialogue. This comparison highlights that the choice of value representation is critical: $m_t^{*}$ serves effectively as a relevance key, whereas $Q_t^{cls}$ is better suited as the recall value for preserving round-specific semantics.

\begin{figure}[tb!]
\centering
\subfloat[]{\includegraphics[width=0.49\linewidth]{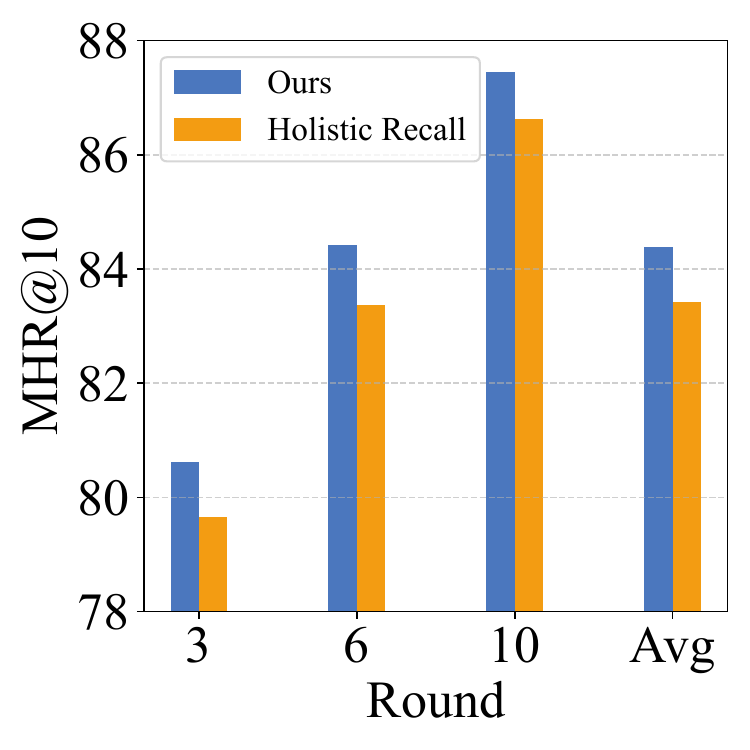}}
\subfloat[]{\includegraphics[width=0.49\linewidth]{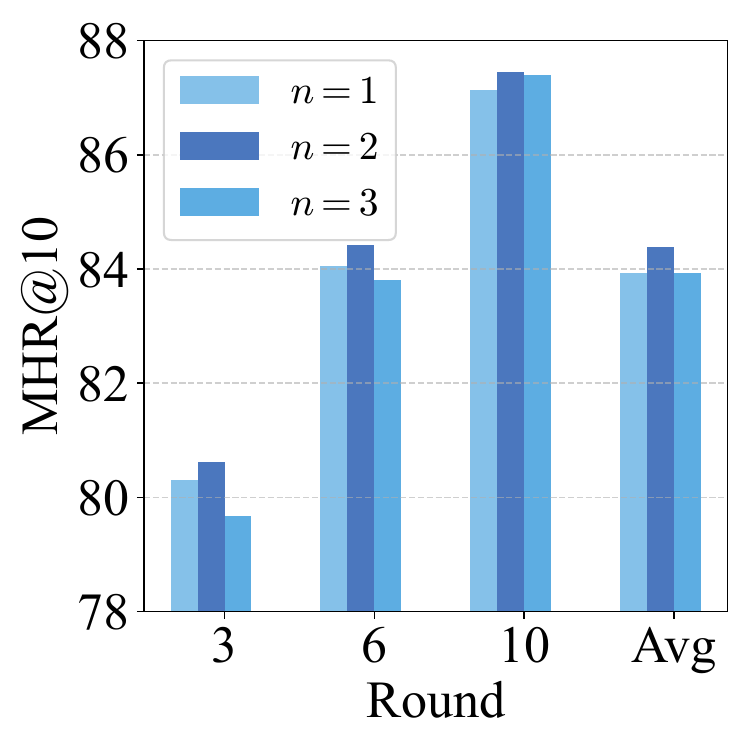}}
\caption{Evaluation of memory recall designs.  (a) similarity-based recall outperforms holistic recall, and (b) a moderate recall scope (two histories) yields the best results .}\label{fig:ablation memory recall}
\end{figure}

\subsubsection{The studies on memory recall}
We firstly compare our similarity-based recall strategy with a holistic baseline that aggregates all previous $t-1$ rounds  using a normalized weighting scheme, where earlier rounds receive higher weights. As illustrated in Fig.~\ref{fig:ablation memory recall} (a), this baseline produces inferior results compared to our method, as it blindly emphasizes the semantics of distant rounds, which may lead to the memory content deviating from the evolving query intent.
In contrast, our similarity-based recall focuses on history that is most complementary to the current memory state, leading to more accurate restoration of forgotten semantics and higher retrieval accuracy across rounds.

We then examine how the number of recalled history units influences performance by retrieving the least similar one, two, or three historical dialogues.
As shown in Fig.~\ref{fig:ablation memory recall} (b), recalling two historical dialogues achieves the best performance. Using only one history unit often under-recovers missing semantics, while recalling three introduces redundant or conflicting information that leads to slight performance degradation.
These results indicate that a moderate recall scope is preferable for effective memory recall.

\begin{figure}[tb!]
\centering
\subfloat[]{\includegraphics[width=0.98\linewidth]{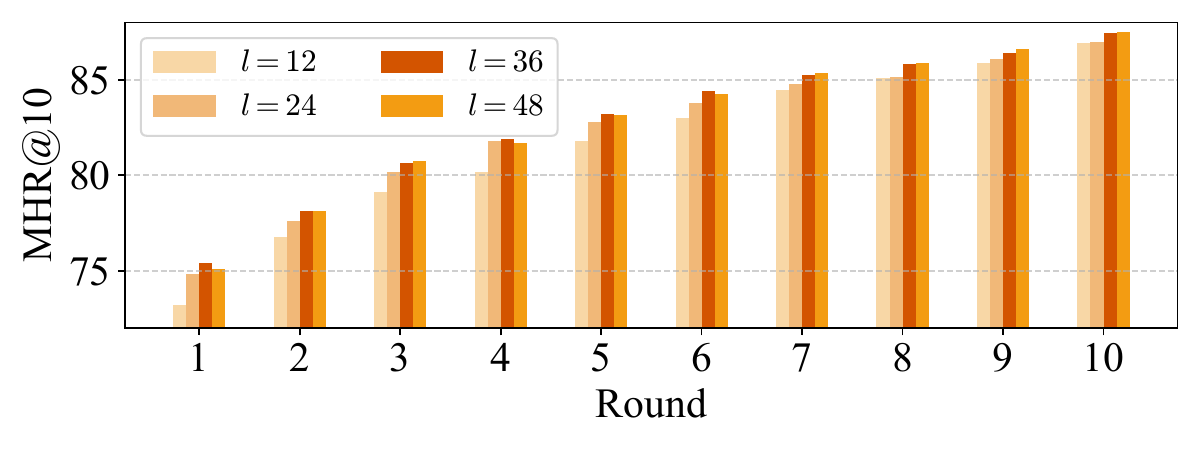}}
\\
\subfloat[]{\includegraphics[width=0.98\linewidth]{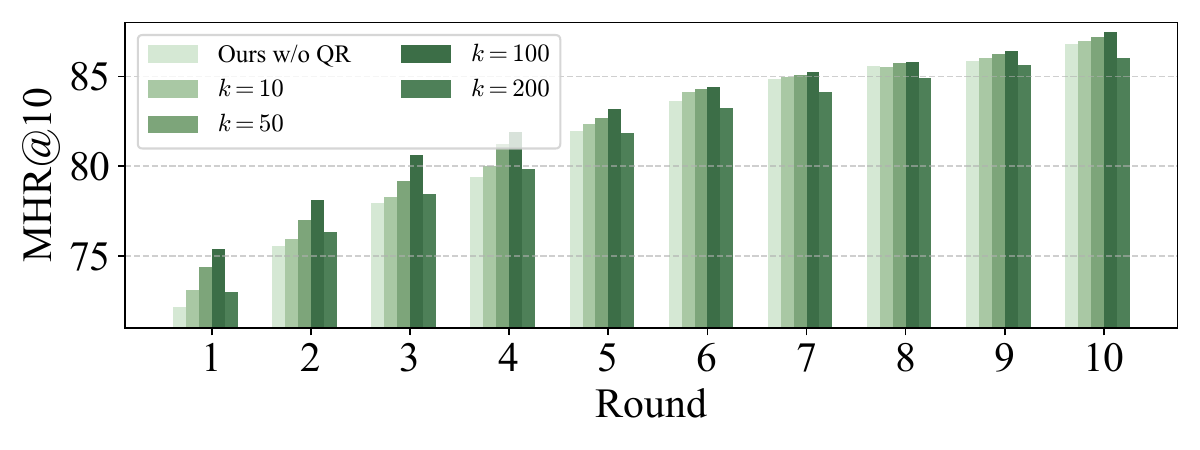}}
\caption{Evaluation of (a) effect of memory token length $l$  and (b) effect of visual history num $k$.}\label{fig:l and k}
\end{figure}



\subsubsection{The Effect of Memory Token Length}\label{sec:token length}
We also investigate how the number of memory tokens $l$ influences the retrieval performance of our method. As shown in Fig.~\ref{fig:l and k} (a), using a smaller token size leads to performance drops across all rounds. This is likely due to insufficient capacity for storing and integrating historical dialogue semantics, resulting in incomplete memory representation. Increasing the token size beyond our default setting does not bring additional performance gains. This suggests that excessively large memory sets introduce redundancy rather than enhancing semantic coverage. Note that
using the number of memory tokens as 36 achieves the best balance of model performance and retrieval cost, so we use it as the default value.

\subsubsection{The Effect of Visual History Number}\label{sec:visual number}

We investigate how the number of retrieved visual candidates $k$ influences the effectiveness of the query refinement process. As shown in Fig.~\ref{fig:l and k} (b), when $k$ is small, the model still achieves better performance than the variant without QR module, since even a limited set of visual cues can provide useful grounding for refining the query intent. However, the restricted candidate pool limits the coverage of relevant visual semantics. Many potentially important visual contexts are not retrieved, which constrains the model’s ability to recover underspecified intent information. As $k$ increases, the available visual evidence becomes more diverse and semantically comprehensive, leading to steady improvements. The performance gains gradually saturate once $k$ becomes sufficiently large, since an overly broad candidate set introduces additional noise, making the refinement process less focused. Overall, $k$=100 provides an effective balance between semantic coverage and noise, and is therefore adopted as the default setting.

\begin{figure*}[tb!]
\centering
\subfloat[]{\includegraphics[width=0.95\linewidth]{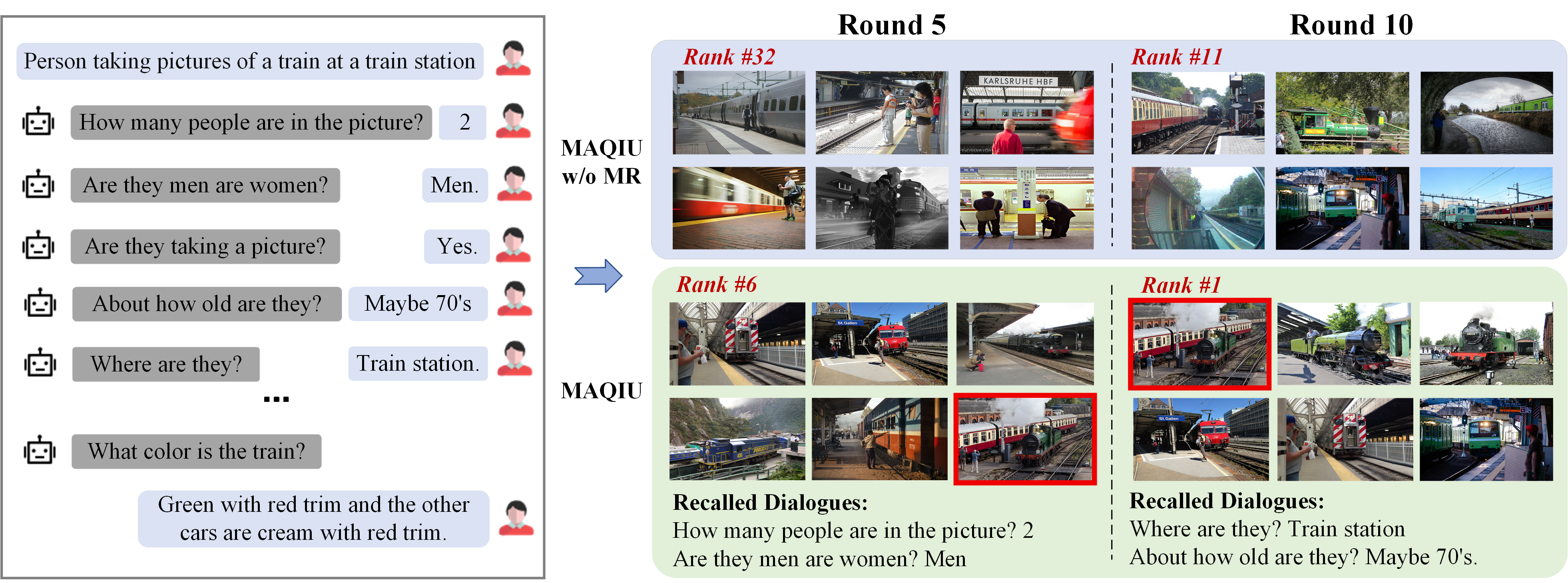}}
\\
\subfloat[]{\includegraphics[width=0.95\linewidth]{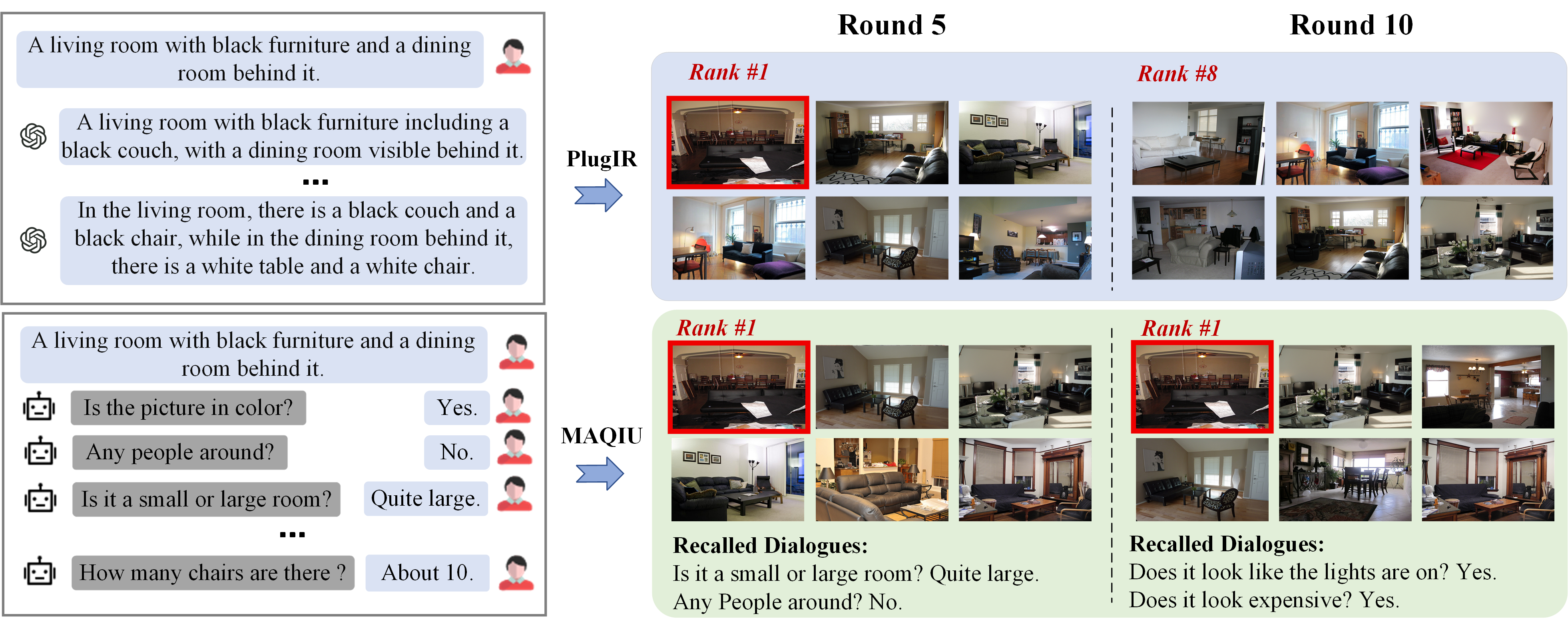}}
\caption{Qualitative multi-round retrieval comparison with different baselines. Each example shows the evolving dialogue on the left, baseline retrieval at rounds 5 and 10 on the top, and our method's retrieval with recalled dialogues on the bottom. (a) Comparison with MAQIU w/o memory recall mechanism. The full MAQIU maintains complete intent representation by recalling forgotten semantics, whereas the counterpart  focuses more on recently dialogue information and suffers from intent drift. (b) Comparison with baseline PlugIR.  Our MAQIU robustly preserves the evolving dialogue intent, whereas baseline approaches exhibit semantic drift and fail to maintain complete intent alignment.}\label{fig:visualize}
\end{figure*}

\subsection{Qualitative Analysis}
Figure~\ref{fig:visualize} (a) provides a qualitative comparison illustrating both the retrieval capability of MAQIU and the contribution of the memory recall mechanism.
At dialogue round 5 and round 10, the full MAQIU retrieves semantically accurate results, while the variant without recall tends to focus narrowly on the most recent dialogue. In particular, when the round-5 dialogue is ``\textit{Where are they? Train station}”, the recall-free model overemphasizes the visual semantic ``train station” and neglects historical semantics, leading to intent drift.
With memory recall enabled, MAQIU retrieves complementary historical information ``two men” that restore forgotten information, thereby constructing a more complete query representation and improving retrieval precision. A similar trend is observed at round 10, further demonstrating that memory recall effectively reinforces long-term semantic completeness across dialogue rounds.

In Fig.~\ref{fig:visualize} (b), although PlugIR produces reasonably correct retrievals at round 5, its performance degrades noticeably at round 10. This degradation arises from the LLM-based reconstruction mechanism. As the dialogue progresses, the reconstruction process may increasingly drifts away from the user’s actual intent. For example, by round 10, the reconstructed query injects irrelevant attributes “a white table and a white chair,” which redirect the model toward mismatched visual concepts. Once these hallucinated details enter the reconstructed text, the retrieval model is forced to follow them, leading to semantic drift and incorrect ranking results. Compared with these baselines, our model maintains coherent intent understanding across rounds. By selectively recalling forgotten semantics and preventing drift, it produces more stable and accurate retrievals throughout the dialogue.

\section{Conclusion}
In this work, we have presented MAQIU, a novel framework for efficient and effective query intent understanding in chat-based image retrieval.
MAQIU maintains a persistent semantic memory to progressively model and refine query intent, while a selective recall mechanism restores forgotten semantics to preserve long-term intent completeness.
Furthermore, a visual feedback strategy enhances cross-round coherence by incorporating historical retrieval results into the current query representation. 